\documentclass{article}

\usepackage[final]{neurips_2025}

\usepackage[utf8]{inputenc} % allow utf-8 input
\usepackage[T1]{fontenc}    % use 8-bit T1 fonts
\usepackage{hyperref}       % hyperlinks
\usepackage{url}            % simple URL typesetting
\usepackage{booktabs}       % professional-quality tables
\usepackage{amsfonts}       % blackboard math symbols
\usepackage{nicefrac}       % compact symbols for 1/2, etc.
\usepackage{microtype}      % microtypography
\usepackage{xcolor}         % colors
\usepackage{float}

\usepackage{amsmath}
\usepackage{amssymb}
\usepackage{mathtools}
\usepackage{amsthm}
\usepackage[table]{xcolor}
\usepackage{colortbl}

\usepackage[normalem]{ulem}

\usepackage{wrapfig}

\usepackage{mathrsfs}
\usepackage[scr=boondoxo]{mathalfa}

\usepackage{soul}

\newcommand{\first}[1]{\cellcolor{green!20}#1}
\newcommand{\second}[1]{\cellcolor{yellow!20}#1}
\newcommand{\third}[1]{\cellcolor{orange!20}#1}

\newcommand{\x}{\mathbf{x}}

\newcommand{\R}{\mathbb{R}}

\newcommand{\hess}[2][]{\mbox{{Hess}}_{#1}\left( #2\right)}
\newcommand{\grad}[2][]{\nabla_{#1}#2}

\newcommand{\fdord}[1]{\left( #1 \right)}
\newcommand{\fdmap}[1]{\left( #1 \right)}

\newcommand{\method}{\text{FLOWING}}
\setcitestyle{square,numbers}

\definecolor{amaranth}{rgb}{0.17, 0.9, 0.31}

\title{FLOWING \includegraphics[scale=0.17]{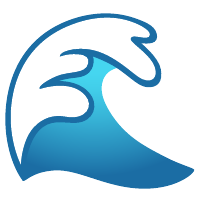}: Implicit Neural Flows\\ for Structure-Preserving Morphing}
\author{%
    Arthur Bizzi$^{1}$ \quad Matias Grynberg$^2$ \quad Vitor Matias$^3$ \quad Daniel Perazzo$^4$\quad \textbf{João Paulo Lima}$^{5,4}$\\[0.2cm]  \textbf{Luiz Velho}$^4$ \quad \textbf{Nuno Gonçalves}$^{6,7}$ \quad \textbf{João Pereira}$^8$ \quad \textbf{Guilherme Schardong}$^6$ \quad \textbf{Tiago Novello}$^4$ \\[0.2cm]
    $^{1}$\small{EPFL} \quad $^{2}$\small{University of Buenos Aires} \quad $^{3}$\small{University of São Paulo} \quad $^{4}$\small{IMPA} \\ $^{5}$\small{Universidade Federal Rural de Pernambuco} \quad $^{6}$\small{ISR-UC} \quad $^{7}$\small{INCM} \quad $^{8}$\small{University of Georgia} %
}

\begin{document}

\maketitle

\begin{abstract}

Morphing is a long-standing problem in vision and computer graphics, requiring a time-dependent warping for feature alignment and a blending for smooth interpolation. Recently, multilayer perceptrons (MLPs) have been explored as implicit neural representations (INRs) for modeling such deformations, due to their meshlessness and differentiability; however, extracting coherent and accurate morphings from standard MLPs typically relies on costly regularizations, which often lead to unstable training and prevent effective feature alignment.
To overcome these limitations, we propose \method~(\textbf{FLOW} morph\textbf{ING}), a framework that recasts warping as the construction of a differential vector flow, naturally ensuring continuity, invertibility, and temporal coherence by encoding structural flow properties directly into the network architectures. This flow-centric approach yields principled and stable transformations, enabling accurate and structure-preserving morphing of both 2D images and 3D shapes.
Extensive experiments across a range of applications—including face and image morphing, as well as Gaussian Splatting morphing—show that \method{} achieves state-of-the-art morphing quality with faster convergence. Code and pretrained models are available in {\small \url{https://schardong.github.io/flowing}}.

\end{abstract}

\section{Introduction}
Morphing is a long-standing problem in computer vision and graphics~\cite{gomes1999warping, wolberg1998image}, with applications in image editing~\cite{smythe1990meshwarp, beier1992feature}, biometrics~\cite{singh2024facemorphing, grimmer2024ladimofacemorphgeneration}, and 3D shape interpolation~\cite{ishit2022levelset,novello2023neural, sang20254deform}.
The task consists of continuously interpolating between two signals while ensuring that the intermediate representations remain structurally consistent.
Traditionally, morphing is decomposed into two stages: a \textbf{warping} stage, which aligns source and target features over time (ideally through a one-parameter family of diffeomorphisms), and a \textbf{blending} stage, which interpolates between the aligned signals.

\begin{figure}[t]
    \centering
    \includegraphics[width=\textwidth]{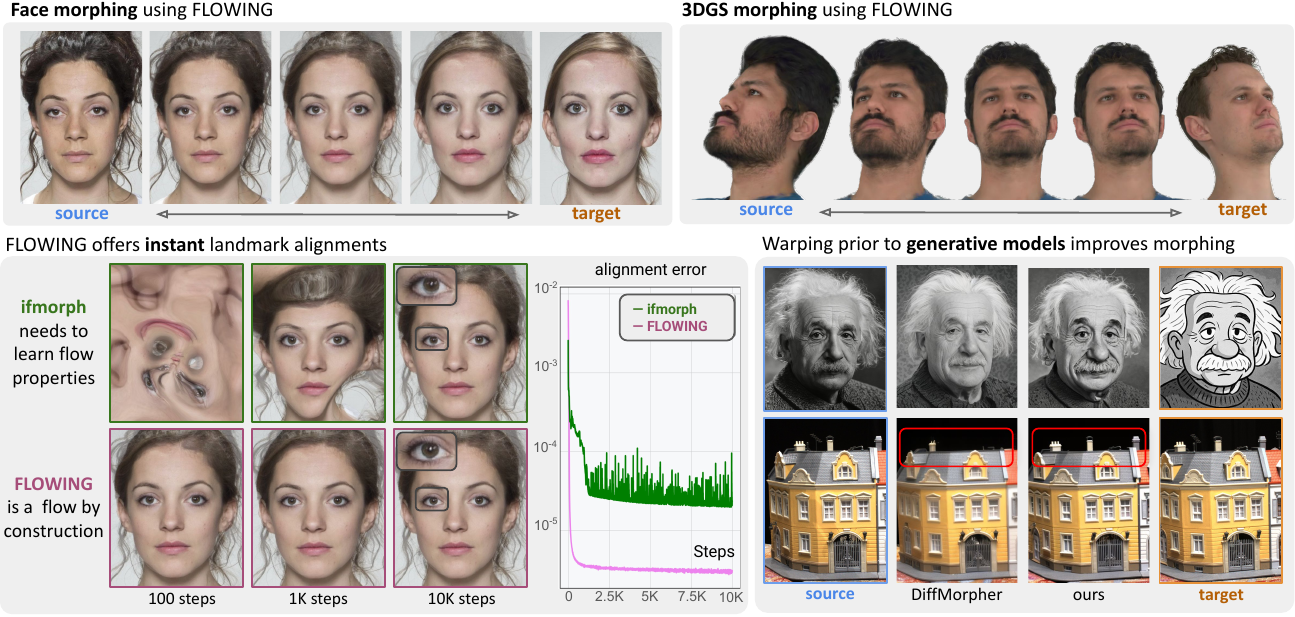}
    \vspace{-0.2cm}
    \caption{We present \textbf{\method}, a robust and theoretically grounded framework for fast, accurate, and structure-preserving morphing.
    It enables smooth and temporally consistent morphs by learning a structure-preserving flow, applicable to both 2D (top-left) and 3D (top-right) data; the latter uses 3DGS for representing 3D faces.
    Bottom-left: \method~instantly aligns landmarks by construction, outperforming SoTA methods in both visual quality and convergence speed.
    Bottom-right: Integrating \method~with generative models improves morphing fidelity and semantic coherence.
    }
    \label{fig:teaser}
\end{figure}

Recently, {ifmorph}~\cite{schardong2024neural} introduced a face morphing approach that employs \textit{multilayer perceptrons} (MLPs) as \textit{implicit neural representations} (INRs) to model the warping transformation. INRs provide a differentiable and memory-efficient solution for representing low-dimensional signals and have been successfully applied to surface reconstruction~\cite{sitzmann2020implicit, gropp2020implicit, yang2021geometry, novello2022exploring, schirmer2024geometric}, surface evolution~\cite{ishit2022levelset, novello2023neural, sang2025implicit}, radiance fields~\cite{mildenhall2020nerf, barron2022mipnerf}, and image modeling~\cite{anokhin2021image, paz2023mr, paz2024spectral}.
However, applying generic MLPs to represent warping transformations presents important limitations. For instance, enforcing temporal coherence requires explicit regularization in the loss function, which greatly increases training time and may lead to convergence issues. In addition, unconstrained MLPs lack structural priors, making the learned deformation prone to undesirable behaviors such as singularities—regions where the mapping becomes non-invertible—resulting in artifacts that degrade the morphing quality (see the first row in \autoref{fig:teaser}, bottom-left).

To address these limitations, we propose \textbf{\method}, a framework that employs specialized flow-based INRs that perform \textbf{structure-preserving warpings by construction}. We reframe morphing as the problem of constructing an interpolating flow, allowing us to inherit the structural properties of flow operators through flow-based neural architectures, such as \textit{neural ODEs} (NODEs)~\cite{chen2018neural} and \textit{neural conjugate flows} (NCFs)~\cite{bizzi2024neural}. \method{} adapts these architectures to the 2D/3D warping contexts and leverages their architectural priors to produce morphings that are valid by construction. As a result, it enables near-instant training, significantly faster than generic MLPs that depend on soft regularization to enforce coherence.
\autoref{fig:teaser} showcases \method{}'s ability to generate smooth, consistent, and structure-preserving morphs in both 2D images and \textit{3D Gaussian splatting} (3DGS)~\cite{kerbl3Dgaussians}, with clear improvements in visual quality, alignment, speed, and semantic coherence over prior methods.
{In summary, our contributions are:}
\begin{itemize}
    \item A principled, flow-based morphing approach that leverages specialized INRs to enforce continuity and temporal coherence by construction. This enables learning highly detailed, structure-preserving morphings near instantly, using only sparse landmark correspondences.

    \item An adaptation of flow-based architectures to the morphing setting, combined with SIRENs~\cite{sitzmann2020implicit, novello2025tuningfrequenciesrobusttraining}, allowing the representation of complex, high-frequency deformations and accurate keypoint matching.
    To minimize spurious deformations, we incorporate thin-plate energy regularization~\cite{bookstein1989principal,olga2007arap}, enabled by a novel forward-mode differentiation scheme. 

    \item Demonstration of our approach across diverse morphing tasks—including face morphing, general image morphing, and 3DGS morphing—achieving strong results in 3D splat morphing via a novel blending scheme.
\end{itemize}

\section{Related works}
\textbf{Warping and morphing} techniques have been widely studied in computer graphics and vision~\cite{gomes1999warping, wolberg1998image, wolberg1990digital}. Classical approaches rely on geometric transformations such as thin-plate splines~\cite{liao2014automating}, radial basis functions~\cite{bookstein1989principal}, and triangle meshes to morph between images~\cite{beier1992feature, olga2007arap}. While effective, these methods often struggle to handle smooth and complex deformations.
Moreover, standard OpenCV-style morphing combines simple linear interpolation between features with a shared triangulated domain, which is not always available. In contrast, by introducing flow concepts into the model architecture, our method (\method{}) achieves high-quality warping for both images and 3D data. It requires only feature correspondences as input and supports nonlinear interpolation between features, enabling more flexible and robust morphing.

\textbf{Implicit neural representations (INRs)} have gained significant attention for encoding continuous low-dimensional signals directly in the parameters of neural networks. Early works such as SIREN~\cite{sitzmann2020implicit} and Fourier feature networks~\cite{tancik2020fourier} enhanced INR expressiveness by mapping input coordinates to sinusoidal functions, enabling detailed reconstructions of images, 3D shapes, and physical fields. More recently, INRs have been applied to 2D face morphing~\cite{schardong2024neural}, where SIRENs are used to parameterize the warping and additional costly constraints are introduced to enforce flow properties.
In contrast, our method leverages flow-based architectures that enable faster, more efficient, and robust morphing of both 2D and 3D data, with deformations learned as continuous, structure-preserving flows.

\textbf{Flow-based networks} have been explored for modeling continuous-time dynamics and transformations. NODEs~\cite{chen2018neural} provide a powerful yet computationally intensive framework for learning continuous-time dynamical systems, where inference through discrete layers is replaced by numerical integration. 
Originally developed for continuous normalizing flows, NODEs have since been applied to medical image registration~\cite{balakrishnan2019voxelmorph, wu2022nodeo, sun2022topology, sun2024medical}, solving PDEs~\cite{bizzi2025neuro}. More recently, NCFs~\cite{bizzi2024neural} were introduced as an alternative approach to model dynamics by topologically deforming affine flows, enabling greater efficiency through parallelism.
In this work, we adapt both NODEs and NCFs to the morphing setting for images and 3DGS, further enhancing their representational power with sinusoidal activations.

\textbf{Generative methods} have also emerged as a powerful and flexible approach for morphing, with DiffMorpher~\cite{zhang2024diffmorpher} presenting general image morphing without reference landmarks. %~\cite{karras2020stylegan, karras2021stylegan, preechakul:2022, zhang2024diffmorpher}.
However, they face two key limitations: their outputs are constrained to a fixed resolution, and they require pre-aligned target images, adding significant preprocessing overhead. Our method addresses these challenges by introducing flow-based warping, which acts as a sophisticated, non-linear alignment mechanism. This enables seamless integration with generative blending techniques, improving both alignment quality and applicability.

%=====================================

\section{\method{}}
\subsection{Problem setup: morphing as flows}
% This may be expressed as follows.
Given source and target media objects \( f^0, f^1: \Omega \subset \mathbb{R}^n \to \mathbb{R}^m \), where $\Omega$ denotes their supports, annotated with $K$ feature correspondences \( \{p^0_i, p^1_i\} \subset \Omega \), our goal is to construct a continuous, time-dependent family of warpings \( \Phi: \Omega \times [0,1] \to \mathbb{R}^n \) that allows us to smoothly interpolate between the source features at \( t = 0 \) and the corresponding target features at \( t = 1 \), such that \( \Phi(p_i^0, 1) = p_i^1 \).
For morphing, $\Phi$ should have additional properties:

$\bullet$ \textbf{Uniqueness.} The path traced by each feature must be {unique}, otherwise distinct features may overlap, leading to {unstructured, incoherent} interpolations.

$\bullet$ \textbf{Invertibility.} The forward and backward mappings of source and target features should match at intermediate times and remain symmetric under time reversal to avoid artifacts during blending.

$\bullet$ \textbf{Energy-minimization.} Feature paths should be minimal and smooth for coherent interpolation; otherwise, overfitting may introduce artifacts or singularities that degrade morphing~quality.

\begin{figure}[h!]
    \centering
    \includegraphics[width=0.87\columnwidth]{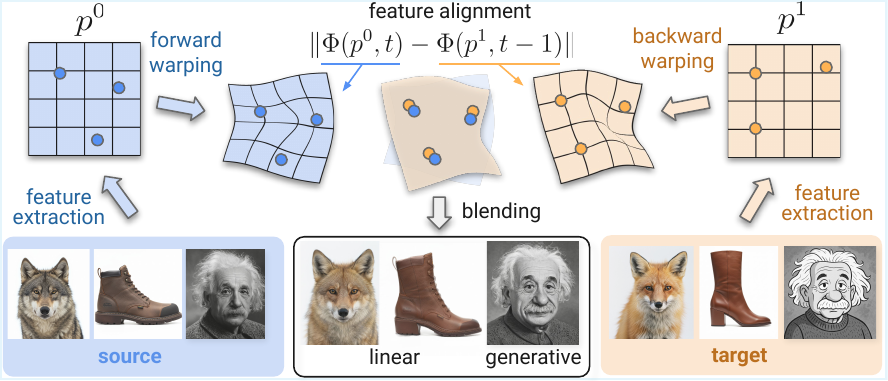}
    \caption{\textbf{Overview of \method{}.}
        Given source and target images $I^0 , I^1$ we extract landmark pairs $(p^0_i,p^1_i) $ with a feature extractor (\texttt{dlib}, Xfeat, etc). 
        % This collection represents the initial and final state for each landmark pair.
        We train a flow $\phi$ such that $\phi(p^0,t) = \phi(p^1,t-1)$, effectively mapping $p^0$ to $p^1$. At inference,
        % to blend the images at time $t$,
        we warp $I^0$ forward by $t$ units and $I^1$ and backward by $t-1$ units,
        % The warped images are then combined using blending methods such as linear blending or generative models.}
    then blend them together with methods such as linear blending or generative models.}
\label{f-overview}
\end{figure}

The key idea behind \method~is that the first two properties arise from the definition of \textbf{flows}: by constraining features to move as an integral over an underlying vector field, we may leverage the uniqueness and reversibility of its orbits. This allows us to recast the morphing problem as the construction of a flow operator \cite{viana2021differential}, enforcing corresponding features to remain consistent over time:
\begin{equation}
        \Phi(p_i^0,t) = \Phi(p_i^1,t-1) \, \text{ for } \,i\in\{1,\ldots, K\} \,\text{ and }\, t\in[0,1].
\end{equation}
To enforce these structural properties on the warping map, we employ a $\theta$-parametrized flow representation $\Phi_\theta$ (see Sec. \ref{s-flow-based-arch}). 
The third property is imposed by minimizing the curvature of the orbits of $\Phi_\theta$ across the domain $\Omega$. This is achieved with the following thin-plate-like optimization problem:
\begin{align}\label{eq:opt-problem}
\displaystyle
\text{arg\,min}_\theta \, \| \mathbf{J}{\Phi_\theta'}(\mathbf{x}, 0) \|_2^2
\quad \text{subject to} \quad \Phi_\theta(p_i^0,t) = \Phi_\theta(p_i^1,t-1),
\end{align}
where $\Phi_\theta'(\mathbf{x}, 0)$ denotes the time derivative of $\Phi_\theta(\mathbf{x}, t)$ at $t=0$ and $\mathbf{J}{\Phi_\theta'}$ its Jacobian. Once the features \( \{p^0_i, p^1_i\}\) are aligned, the resulting flow can be used to warp the source and target signals to an intermediate time, after which the resulting warpings are blended together. \autoref{f-overview} provides an overview of the morphing process.

\subsection{Training}\label{sec:training}

Training a flow-based architecture $\Phi_\theta$ requires defining a loss function to solve the optimization problem in \eqref{eq:opt-problem}. Note that the uniqueness and invertibility properties are guaranteed by construction, since $\Phi_\theta$ is a flow.
Therefore, the loss function $\mathcal{L}$ consists of a data constraint to enforce feature matching and a regularization term to penalize path distortion:
\begin{align}\label{eq:loss-flow}
\mathcal{L}(\theta) =
\underbrace{\sum_i \int_{[0,1]} \left\| \Phi_\theta(p^0_i,t) - \Phi_\theta(p^1_i,t-1) \right\|_2^2\,dt}_{\text{Data constraint}}
+ \underbrace{\lambda \int_\Omega \left\| \mathbf{J}{\Phi_\theta'}(\mathbf{x}, 0) \right\|_2^2 \, d\mathbf{x}}_{\text{Thin-plate constraint}},
\end{align}
where $\lambda>0$ is a parameter.
In practice, the data term is enforced on only a few time steps, since uniqueness ensures that feature correspondences hold across the interval. We find out that using $t=0.5$ for NODE and $t=\{0,0.25,0.5,0.75,1\}$ for NCF is sufficient.
% (\tiago{@matias, see ablation on training time steps in ???)}.

For thin-plate term, we minimize the Jacobian of ${\Phi_\theta'}(\mathbf{x}, 0)$ over the domain, which is equivalent to minimizing the second derivative of the integral path $\Phi(\mathbf{x}, t)$ for each $\mathbf{x}$. Since $\Phi$ is a flow, explicit time sampling is unnecessary—one reason flow-based architectures train faster than generic MLPs. The integral is approximated via Monte Carlo methods, with $\lambda$ as the regularization weight.
To further accelerate computation, we implement a \textit{forward differentiation} (FD) scheme based on generalized dual-number arithmetic, enabling derivatives to be computed in parallel with inference, significantly reducing overhead. Our ablation study shows that FD makes the thin-plate loss calculation \textbf{23 times faster} than standard \texttt{autograd}. Full results and details are given in \autoref{a-FD}.

\subsection{Flow-based architectures}\label{s-flow-based-arch}
Flow-based networks provide a principled representation for the time-dependent warping $\Phi_\theta$, as they hold desirable properties such as continuity, invertibility, and temporal coherence. 
Compared to using MLPs to parametrize the flow operator $\Phi_\theta$, which require expensive additional terms to approximate uniqueness and invertibility in addition to (\ref{eq:opt-problem}), flow-based architectures enforce these properties by construction. As a result, they rely on simpler losses, may achieve near-instant training, and provide accurate alignment while avoiding the catastrophic artifacts that arise when MLPs fail to capture proper flow properties. 

\begin{wrapfigure}[16]{R}{0.57\textwidth}
    \centering
    % \vspace{-0.15cm}
    \includegraphics[width=0.94\linewidth]{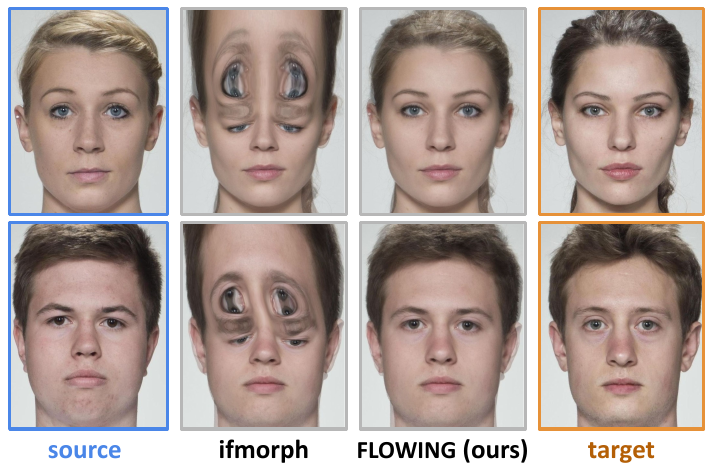}
    \vspace{-0.05cm}
    \caption{Comparison between {ifmorph} and \method{}. While ifmorph fails to preserve structure, \method{} produces clean and realistic interpolations by enforcing flow properties such as invertibility and trajectory uniqueness.}
    \label{fig:catastrophic-morphs}
\end{wrapfigure}
\autoref{fig:catastrophic-morphs} compares an MLP-based~flow (ifmorph~\cite{schardong2024neural}) with our approach (\method{}). For this experiment, we employed \texttt{dlib} to extract features centered on the faces, which explains why the ears are not aligned. In this work, we focus on two flow-based networks: Neural ODEs and Neural Conjugate Flows.

\paragraph{Neural ODEs.}
First, we propose using NODEs~\cite{chen2018neural} to model the vector field $\Phi'$, the time derivative of the warping deformation $\Phi$, with a neural network $\mathcal{F}_\theta$. Thus, inference for these models involves performing numerical integration over the vector field $\mathcal{F}_\theta$; analogously, it may be interpreted as deep ResNets\cite{he2016resnet} with ``continuous depth''. Specifically, NODEs take the following formulation:
\begin{equation}\label{e-node}
    \begin{aligned}
        \frac{d}{dt}\x &= \mathcal{F}_\theta(\x) \implies \\ \Phi(\x_0,t) &= \x(0) \!+ \!\!\int_0^t\!\! \mathcal{F}_\theta\big(\x(\tau)\big)d\tau.
    \end{aligned}
\end{equation}
To achieve high-quality matching, the vector field $\mathcal{F}_\theta$ is modeled as a SIREN~\cite{sitzmann2020implicit}, allowing it to capture highly detailed spatial deformations necessary to handle the diverse range of distances and orbits traversed by each pair of correspondences. Moreover, to ensure that orbit deformations remain minimal, we penalize the norm of the Jacobian matrix of $\mathcal{F}_\theta$ over the domain $\Omega$.

During training, we integrate $\mathcal{F}_\theta$ using a fourth-order Runge–Kutta method. We integrate the source and target points to $t=0.5$ using the same number of integration steps for the forward and backward dynamics. Since NODE-based approaches are often prone to numerical errors during integration, we include in \autoref{a-ablation} an ablation study on the total number of integration steps involved in this process, showing that we can use very few integration steps while both preserving an accurate approximation of the vector field $\mathcal{F}_\theta$ and keeping the training efficient. Moreover, at inference time, the number of integration steps can be increased arbitrarily, without additional training cost, to more accurately capture the dynamics of $\mathcal{F}_\theta$.

\paragraph{Neural conjugate flows.}
While NODEs explicitly integrate neural vector fields, they can become computationally expensive when many steps are required during inference. Neural conjugate flows (NCFs)~\cite{bizzi2024neural} offer an alternative formulation based on topological conjugation. Instead of sequential integration, NCFs employ invertible neural networks to deform the orbits of simple affine flows, enforcing a simplified topology. 
Precisely, the flow-based warping is given by:
\begin{equation}
   \Phi_\theta(\x,t) = H_\theta^{-1} \circ \Psi\big(H_\theta(\x),t\big), \quad (\x,t) \in \mathbb{R}^n \times \mathbb{R}.
\end{equation}
We parameterize $H_\theta$ and $H_\theta^{-1}$ using invertible architectures known as coupling layers~\cite{dinh2017density}. These apply alternating, memory-equipped transformations in sequence, guaranteeing invertibility by construction. Importantly, the resulting conjugated flow $\Phi_\theta$ remains a valid flow, inheriting key properties such as continuity, invertibility, and associativity from the original affine system. 
% \joaopaulo{wouldn't it be better to move the next paragraph to the intro of Subsection 3.3? now it is in the "NCF" subsubsection and it mentions ``near-instant training'', which I believe is more suitable to NODE, no?}

\subsection{Blending}

Having established how \method{} constructs flow-based warpings that guarantee structural properties and accurate feature alignment, we now turn to the second stage of morphing: \textbf{blending}.
\paragraph{Image morphing.}
Let $I^0,\,I^1 : \R^2 \to \mathcal{C}$ denote two input images and $\Phi_\theta : \R^2 \times \R \to \R^2$ be a flow aligning their features over time. We define the warped images as $\mathscr{I}^i(\x, t) = I^i\big(\Phi_\theta(\x, i - t)\big)$, ensuring spatial alignment at each intermediate timestep $t \in [0, 1]$.
A straightforward morphing is then obtained by linearly blending the aligned images:
\begin{align}
\mathscr{I}(\x, t) = (1 - t)\mathscr{I}^0(\x, t) + t\mathscr{I}^1(\x, t),
\end{align}
yielding a smooth function $\mathscr{I} : \R^2 \times \R \to \mathcal{C}$ that interpolates between $I^0$ and $I^1$.

While generative models can interpolate between images, they typically lack explicit feature alignment. 
To address this limitation, we first apply our flow-based warping $\Phi_\theta$ to align features over time, and then perform blending in the latent space of a pretrained generative model. Let $\mathscr{E}$ and $\mathscr{D}$ denote the encoder and decoder of a generative model. We define generative morphing as:
\begin{align}
    \mathscr{I}(\cdot, t) = \mathscr{D}\left((1 - t)\mathscr{c}^0(t) + t\mathscr{c}^1(t)\right), \quad
    \text{where} \quad
    \mathscr{c}^i(t) = \mathscr{E}\big(\mathscr{I}^i(\cdot, t)\big).
\end{align}
This strategy combines explicit spatial alignment with perceptual quality, producing temporally coherent transitions that significantly outperform naive latent-space blending.

\paragraph{3D Gaussian Splatting morphing.}
Beyond images, we show that morphing with \method{} can be extended to 3DGS~\cite{kerbl3Dgaussians, matias2025gaussian} which has recently emerged as a popular format for 3D representation. 3DGS enables photorealistic rendering from a set of Gaussians $\mathscr{G}$ with each Gaussian $\mathscr{g}_k = \{p_k, \alpha_k, c_k, \Sigma_k\}$ consisting of a center $p_k \in \R^3$, opacity $\alpha_k$, view-dependent color $c_k \in \R^3$, and covariance matrix $\Sigma_k \in \R^{3\times 3}$ that encodes orientation and spread.

Our method morphs between two Gaussian sets, $\mathscr{G}^0$ and $\mathscr{G}^1$, producing an intermediate set $\mathscr{G}(t)$ with $\mathscr{G}(0)=\mathscr{G}^0$ and $\mathscr{G}(1)=\mathscr{G}^1$. We employ a \method{} network $\Phi_\theta$ to smoothly align Gaussian centers over time, while linearly blending the opacity parameters $\alpha_k^i$ to ensure seamless transitions. Specifically, for each Gaussian $\mathscr{g}_k^i \in \mathscr{G}^i$, we compute:
\begin{equation}\label{eq:gau_blend} \mathscr{g}_k^i(t) = \bigl\{\Phi_\theta(p_k^i,\;t-i),\;|1-i-t|\,\alpha_k^i,\;c_k^i,\;\Sigma_k^i\bigr\}, \end{equation}
which jointly applies warping and blending. The final morphed set is then defined by $\mathscr{G}(t) = \mathscr{G}^0(t)\cup\mathscr{G}^1(t)$.
This {3DGS morphing} procedure preserves structural coherence through flow-based alignment and achieves photorealistic 3D transitions via our linear blending strategy.

\section{Experiments}

We evaluate \method{} on face and image morphing, as well as on 3DGS morphing, using four diverse datasets. These experiments demonstrate the effectiveness of our approach in both 2D and 3D settings. 
Additionally, we provide ablation studies in \autoref{a-ablation} to validate our architectural choices and regularization strategies.

\textbf{Evaluation datasets.}
For face images, we use the FRLL dataset~\cite{debruine2017frll}, which contains $102$ identities with $5$ images captured at fixed angles and two expressions (neutral and smiling). We use the neutral, front-facing images. Following the protocol in \cite{sarkar2020landmarks}, we construct $1220$ pairs, with landmarks extracted using \texttt{dlib}~\cite{kazemi2014cvpr, sagonas2016}. 
As a post-processing step, we apply FFHQ alignment and cropping, producing images at a resolution of $1350^2$.
The second dataset is a subset of MegaDepth~\cite{li2018megadepth}, which provides multi-view scene landmarks. From this subset, we extract $275$ image pairs based on the cosine similarity between their ResNet~\cite{he2016resnet} embeddings, with feature correspondences obtained using Xfeat~\cite{potje2024xfeat}.
Third, we use eight subjects from NeRSemble~\cite{kirschstein2023nersemble}, a multi-view collection of human heads. Following \cite{qian2024gaussianavatarsphotorealisticheadavatars}, we fit an FLAME model to each subject to extract $254$ landmarks per head.
Finally, we include in-the-wild face images from FFHQ~\cite{karras2020stylegan} for qualitative evaluation.

% \paragraph{Training details.}
\method~is implemented in PyTorch~\cite{paszke2019pytorch} and trained with the Adam optimizer~\cite{kingma2014adam}.
%Unless otherwise specified, experiments are conducted by evaluating the loss function~\eqref{eq:loss-flow} on the set of feature pairs at five equally spaced values of $t \in [0, 1]$.
At each training step, we sample $20{,}000$ points from the spatial domain $[-1, 1]^2$ for the selected values of $t$.

\subsection{Quantitative comparisons} 
% \joaopaulo{``Quantitative Comparisons" would be a better subsection title? Because we also do comparisons in Subsections 4.2 and 4.3}
To evaluate the performance of \method{} on the morphing task, we consider two key components: \textbf{warping}, which aligns landmarks across signals over time, and \textbf{blending}, which generates smooth intermediate transitions. Accordingly, we organize our comparisons into two aspects: \textit{landmark alignment} (\ref{sec:results-la}) and \textit{blending quality} (\ref{sec:results-blend}).

\subsubsection{Landmark alignment}
\label{sec:results-la}

We compare \method{} with ifmorph~\cite{schardong2024neural}, testing both NCF and NODE backbones with sigmoid and SIREN activations. 
For a fair comparison, all methods are supervised under the same set of landmark points.

\textbf{Metrics.}
We report the \textit{mean squared error} (MSE) between warped target landmark positions across intermediate timesteps. 
For each landmark pair, we sample $10$ evenly spaced values of $t \in [0, 1]$, warp the landmarks accordingly, and compute their MSE, averaging the results across all timesteps. 
A perfect alignment would yield an average MSE of $0$, since source and target landmarks would coincide exactly at every value of $t$.
Experiments are conducted on three benchmarks: face / monument / 3D Gaussian avatar alignment. 

% \paragraph{Model Size.}
% To contextualize performance, we also report the number of parameters used by each method. \method~requires significantly fewer parameters than IFMorph while achieving comparable or better alignment.\tiago{should we keep this?} \guilherme{I'm not sure. I would remove it, since even with fewer parameters, NCF takes longer to train, unless its orders of magnitude of difference.}

{\textbf{Sinusoidal vs. non-sinusoidal activations.}
We compare NCF- and NODE-based variants of FLOWING using SIREN activations and their sigmoid counterparts to justify our architectural choices for high-frequency representations in morphing quality and convergence.
\autoref{t-sota} reports the average MSE across the FRLL, MegaDepth, and NeRSemble datasets for face warping, monument alignment, and 3D Gaussian avatar morphing.
Results are shown for ifmorph, NCF, and NODE. For NCF and NODE, we evaluate both SIREN and sigmoid activations.
Note that NCF (SIREN) and NODE (SIREN) outperform ifmorph by one to three orders of magnitude, while SIREN activations consistently yield lower errors than the sigmoid case. In particular, NODEs with sigmoid (“vanilla” configurations) perform significantly worse, confirming the importance of sinusoidal activations for accurate warping alignment.
\begin{table}[!h]
% \small
\centering
\caption{Results for the landmark alignment across the FRLL~\cite{debruine2017frll}, MegaDepth~\cite{li2018megadepth} and NeRSemble~\cite{kirschstein2023nersemble} datasets. The best, second and third best results for each dataset/model combination are shown in \textcolor{green!60!black}{green}, \textcolor{yellow!70!black}{yellow}, and \textcolor{orange!80!black}{orange}, respectively.}
\label{t-sota}
% \small
\begin{tabular}{lrrr}
\toprule
Model (Activation) & FRLL~($\downarrow$) & MegaDepth~($\downarrow$) & NeRSemble~($\downarrow$) \\
\midrule
ifmorph~\cite{schardong2024neural}           & \third{1.5E-3}    & 2.9E-1            & 1.0E-3         \\
NCF (sigmoid)     & 8.7E-3            & 2.0E+1            & \second{7.2E-5} \\
NCF (SIREN)       & \first{3.9E-5}    & \first{4.2E-4}    & \third{7.3E-5} \\
NODE (sigmoid)    & 6.2E-2            & \third{}{1.9E-1}  & 5.0E-4         \\
NODE (SIREN)      & \second{1.4E-4}   & \second{}{4.9E-4} & \first{5.9E-5} \\
\bottomrule
\end{tabular}
\end{table}

\textbf{NCF vs. NODE.}
\autoref{fig:trainning_run} shows that NCFs and NODEs achieve better results with significantly fewer training steps compared to ifmorph. While NODEs converge quickly, they typically show poorer feature alignment. In contrast, NCFs require more steps to converge, resulting in better feature alignment overall. However, due to minor pixel-level variations between source and target features, this does not necessarily translate to noticeable visual differences.
\begin{figure}[!h]
    \centering
    \includegraphics[width=\textwidth]{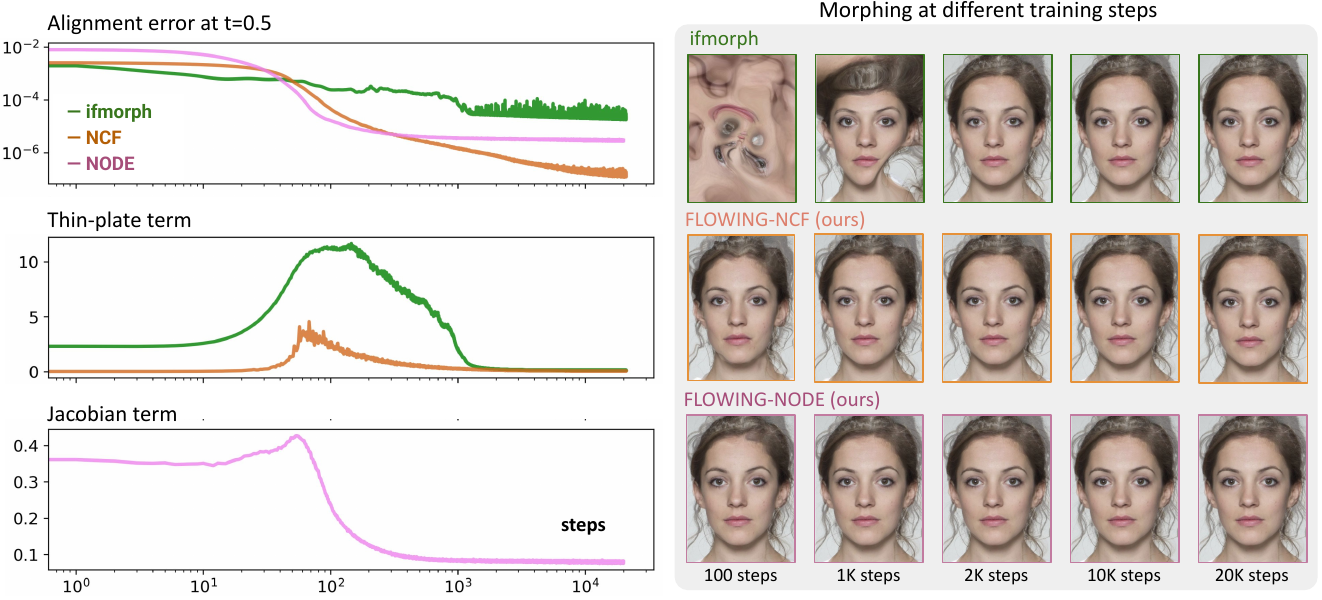}
    \vspace{-0.2cm}
   \caption{Convergence analysis of ifmorph (\textcolor{green!60!black}{green}), NCF (\textcolor{orange!80!black}{orange}), and NODE (\textcolor{violet!80!black}{violet}). Left: alignment and deformation metrics over training steps. 
For NODE, alignment is measured using the Jacobian term, while ifmorph and NCF use the Hessian-based thin-plate term. 
Right: qualitative morphing results at $t=0.5$ after $100$, $1000$, $2000$, $10000$, and $20000$ training steps. 
NCF and NODE achieve accurate alignment by $1000$ steps, whereas ifmorph requires at least $2000$ steps.}
    \label{fig:trainning_run}
\end{figure}

% \begin{table}[h]
% \centering
% \caption{Warp training times for different methods with 68 and 130 landmarks.}
% \begin{tabular}{lccc}
% \toprule
% {Model} & {Steps} & {Warp time (68)} & {Warp time (130)} \\
% \midrule
% ifmorph     & 20k  & 05m26s & 05m22s \\
% NCF (Ours)  & 20k  & 08m48s & 15m05s \\
% NODE (Ours) & 2k   & 00m16s & 00m15s \\
% NODE (Ours) & 20k  & 02m39s & 02m38s \\
% \bottomrule
% \end{tabular}
% \end{table}

% \begin{table}[h]
% \caption{Morphing times (in seconds) for different methods at resolutions 256×256 and 1350×1350.}
% \centering
% \begin{tabular}{lcc}
% \toprule
% \textbf{Method} & \textbf{Morphing time (256×256)} & \textbf{Morphing time (1350×1350)} \\
% \midrule
% OpenCV      & 0.01 & 0.06 \\
% ifmorph     & 0.05 & 0.10 \\
% NCF (Ours)  & 0.02 & 0.31 \\
% NODE (Ours) & 0.03 & 1.48 \\
% \bottomrule
% \end{tabular}
% \end{table}

\subsubsection{Blending quality}
\label{sec:results-blend}
We evaluate blending quality using FRLL~\cite{debruine2017frll} for linear face morphing, following the same experimental setup as in the landmark alignment experiment (Sec.~\ref{sec:results-la}).
We compare \method{} with SIREN activations against {ifmorph} and a standard OpenCV baseline using Delaunay triangulation warping with linear blending\footnote{\url{https://github.com/Azmarie/Face-Morphing/}}.
Additional comparisons with the classical thin-plate spline method are given in~\autoref{a-tps}, and further experiments on video interpolation are discussed in~\autoref{a-video}.

\textbf{Metrics.} Perceptual quality is measured using the \textit{learned perceptual image patch similarity} (LPIPS) metric~\cite{zhang2018perceptual}, computed between intermediate frames at $t = 0.5$ and both source and target images.
We also measure the \textit{Fréchet inception distance} (FID)~\cite{heusel2017fid} between generated morphs and original images, using 2048-dimensional feature vectors.

% \textbf{Results.}
\begin{table}[!h]
\centering
% \small
\caption{Linear image blending results on the FRLL dataset~\cite{debruine2017frll}. The best, second-best, and third-best results are highlighted in \textcolor{green!60!black}{green}, \textcolor{yellow!70!black}{yellow}, and \textcolor{orange!80!black}{orange}, respectively.}
\label{t-blending}
% \small
\begin{tabular}{@{}llll@{}}
\toprule
Morphing type & \multicolumn{1}{l}{LPIPS$(I^0,I)~(\downarrow)$} & \multicolumn{1}{l}{LPIPS$(I,I^1)~(\downarrow)$} & \multicolumn{1}{l}{FID$~(\downarrow)$} \\ \midrule
% TPS Warping  & 0.371 & 0.374 & 124.55 \matias{Enormous! Should we leave this?} \\
OpenCV        & \third{0.233}                                        & \third{0.236}                                        & \second{32.426}                              \\
ifmorph~\cite{schardong2024neural}       & 0.250                                        & 0.252                                        & 38.427                              \\
NCF (Ours)           & \second{0.221}                                        & \second{0.224}                                        & \third{33.300}                              \\
NODE (Ours)          & \first{0.210}                                        & \first{0.213}                                        & \first{31.952}                              \\ \bottomrule
\end{tabular}
\end{table}

\begin{wrapfigure}[30]{R}{0.52\textwidth}
    \centering
    \vspace{-0.5cm}
    \includegraphics[width=\linewidth]{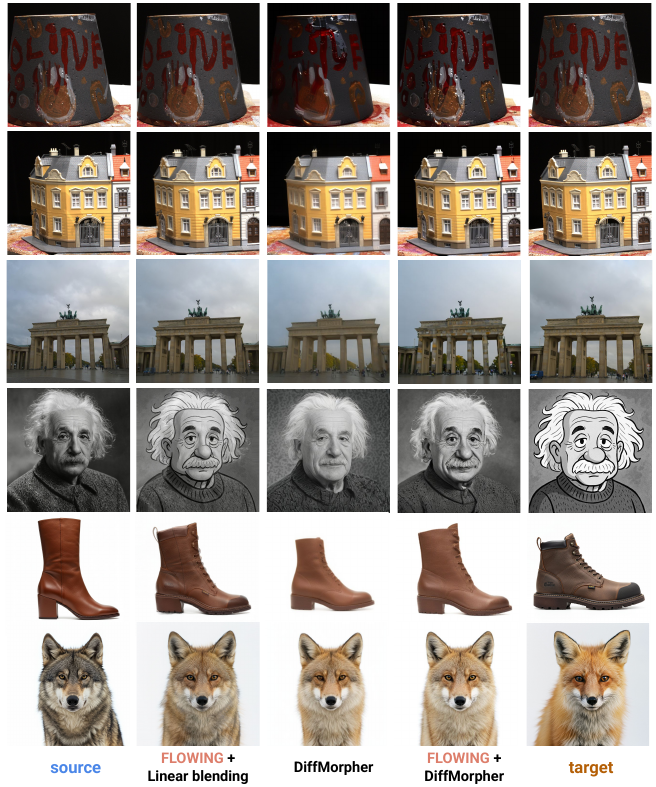}
    \vspace{-0.5cm}
    \caption{Applications of \method{} to {view interpolation} (Rows 1-3), and {stylization and object morphing} (Rows 4-6). \method{} produces smooth transitions even under environmental variations, maintaining geometric and photometric consistency. When combined with DiffMorpher, it further enhances structural coherence and visual fidelity, recovering details missed by DiffMorpher alone, such as the missing chimneys and the loss of column detail (Rows 2-3).}
    \label{fig:general-morphing-examples}
\end{wrapfigure}
Table~\ref{t-blending} shows that both NCF and NODE outperform {ifmorph} across all metrics.
NCF and NODE also obtained better results than OpenCV in LPIPS.
For FID, NODE achieves the best overall performance, while NCF remains competitive, slightly below OpenCV.
Overall, NODE produces higher-quality and more consistent morphing results than NCF.

\subsubsection{Warping and morphing times} 
% \joaopaulo{I think this should be a subsubsection of Subsection 4.1, since it is a quantitative comparison}
% \joaopaulo{I think this paragraph needs to be a separate subsubsection instead of being inside the "Landmark Alignment" subsubsection, since it also deals with blending} 
\autoref{t:times} summarizes training and morphing times obtained on an RTX~4090 GPU for both warp training and morphing (warping inference + blending). We report NODE with both 2000 and 20000 training steps to highlight its fast convergence. As shown, training times for NODE and ifmorph remain stable across different landmark counts, whereas NCF exhibits increased training time with additional landmarks. Morphing time depends on image resolution: NODE provides faster training but slower morphing, while NCF has slower training and intermediate morphing~speed.

\subsection{Image and face morphing}
As shown in the previous sections, \method{} can be used to warp between two images given a set of landmark correspondences. The warped images can then be blended using various techniques, such as linear blending, Poisson image editing~\cite{perez2023poisson}, or generative blending~\cite{zhang2024diffmorpher}. In this work, we focus primarily on linear blending and generative blending via DiffMorpher~\cite{zhang2024diffmorpher}.
Rows 1–3 of \autoref{fig:general-morphing-examples} illustrate the view interpolation task, where \method{} produces smooth transitions while better preserving both geometric and photometric consistency across viewpoints.
These examples show the benefit of applying \method{} prior to generative blending in contrast to the usage of generative blending on its own. It improves both the consistency of the scene structure and reduces the number of missing elements or details. For instance, the building chimneys (Row 2) and architectural details (Row 3) remain intact, unlike in competing approaches.
Rows 4–6 demonstrate stylization and object morphing, where combining \method{} with generative models enhances structural coherence and visual fidelity across diverse visual domains, for example, in the photo-to-cartoon Einstein interpolation.
\begin{table}[h!]
\centering
\caption{Warp training times (68 and 130 landmarks) and morphing times at resolutions $256^2$ and $1350^2$. NODE is reported for both 2k and 20k training steps to highlight its fast convergence.}
\label{t:times}
% \small
\begin{tabular}{llllll}
\toprule
            &       & \multicolumn{2}{c}{Warping training} & \multicolumn{2}{c}{Morphing time} \\
Method      & Steps & 68 landmarks     & 130 landmarks     & Res.  $256^2$   & Res. $1350^2$   \\ \hline
OpenCV      & --    & --               & --                & 0.01s           & 0.06s           \\
ifmorph~\cite{schardong2024neural}     & 20k   & 05m26s           & 05m22s            & 0.05s           & 0.10s           \\
NCF (Ours)  & 20k   & 08m48s           & 15m05s            & 0.02s           & 0.31s           \\
NODE (Ours) & 2k    & 00m16s           & 00m15s            & 0.03s           & 1.48s           \\
NODE (Ours) & 20k   & 02m39s           & 02m38s            & 0.03s           & 1.48s           \\ \bottomrule
\end{tabular}
\end{table}

\autoref{fig:wild-faces-morphing} shows the application of \method{} to non-aligned, in-the-wild face images.
Using the NODE backbone for warping and DiffMorpher for blending, our method produces morphings that are both visually coherent and perceptually realistic.
Additional examples of generative morphing results are provided in \autoref{a-generative}.
\begin{figure}[!ht]
\centering
\includegraphics[width=\columnwidth]{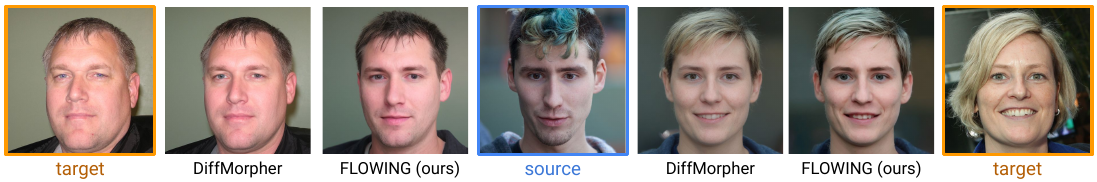}
\vspace{-0.6cm}
\caption{Qualitative results on in-the-wild faces from the FFHQ dataset~\cite{karras2020stylegan}.
By applying our flow-based warping prior to generative blending, \method{} produces morphings that are significantly more coherent and realistic than those generated by DiffMorpher.}
\label{fig:wild-faces-morphing}
\end{figure}

\subsection{3D face morphing using Gaussian splatting}

In this section, we evaluate our 3D face morphing framework using GaussianAvatars~\cite{qian2024gaussianavatarsphotorealisticheadavatars}, an extension of 3DGS for photorealistic human head representation.  
Each face in GaussianAvatars is associated with a FLAME model~\cite{li2017learning}, which enables the extraction of 3D facial landmarks spatially aligned with the Gaussian distribution.  
To morph between two GaussianAvatars, we apply the warping and blending formulation defined in~\eqref{eq:gau_blend}.
\autoref{fig:gaussian_morphing_examples} provides an overview of \method{} applied to 3DGS morphing.  
On the left, we visualize the 3D flow field between two subjects derived from their landmarks and illustrate how the warped Gaussians can be combined to form an intermediate representation at \(t = 0.5\), blending structural and appearance features from both faces.  
The middle panel shows additional morphing results for different subjects across multiple time steps.  
Our method achieves smooth and geometrically consistent transitions between subjects with distinct facial structures and appearance attributes.  
In particular, it naturally handles complex variations such as hair (second and third rows), thanks to the volumetric nature of 3DGS.

Finally, the right panel compares different blending strategies.  
The top row corresponds to a purely 2D setting, where warping is applied to the rendered images followed by linear blending, which results in severe misalignments and ghosting artifacts.  
In contrast, the middle and bottom rows show our 3D warping results, first with linear blending of the rendered images, and then with direct blending of the 3D Gaussians.  
The fully 3D pipeline produces smoother and more coherent transitions, effectively eliminating ghosting and preserving fine structural details across viewpoints.  
Additional qualitative examples of 3D morphing results are presented in \autoref{a-3dgs}.
\begin{figure}[!ht]
    \centering
    \includegraphics[width=\columnwidth]{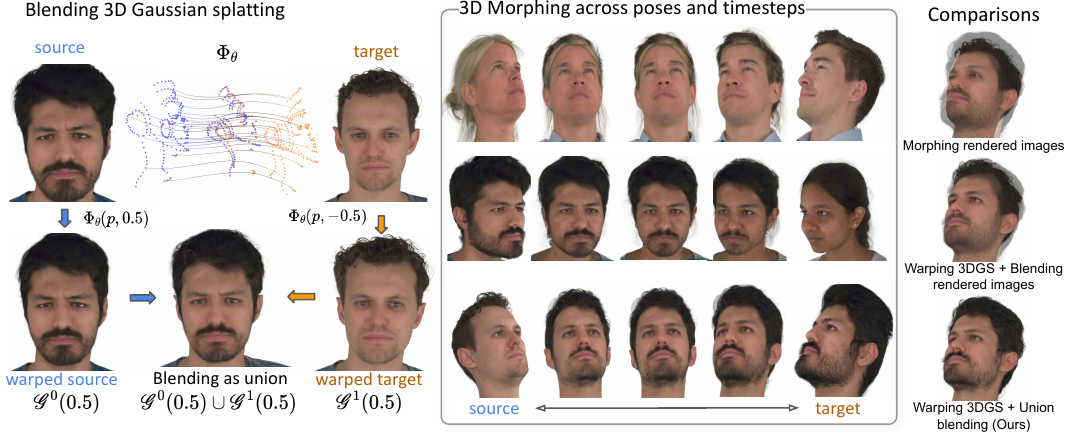}
    \vspace{-0.3cm}
    \caption{Overview of \method{} for morphing between 3D faces using Gaussian splatting.  
    Our approach warps the space to enforce 3D landmark alignment and applies {union-based 3DGS blending}, directly combining Gaussians in 3D space (left).  
    This yields geometrically consistent morphs and preserves photorealistic appearance across poses and viewpoints (middle).  
    The right panel compares blending strategies, showing that our union-based 3DGS blending produces smoother and more coherent results than morphing rendered images or blending after warping.}
    \label{fig:gaussian_morphing_examples}
\end{figure}

\section{Conclusion and limitations}

This work introduced \method{}, a novel framework for morphing between graphical objects using flow-structured INRs.  
By leveraging the mathematical properties of flows, our approach enables faster convergence, more stable training, and robust warping behavior that avoids catastrophic singularities.  
We demonstrated the versatility of \method{} across a wide range of tasks—including face morphing, view interpolation, and Gaussian-based 3D morphing—achieving high-quality and temporally coherent results.  
Moreover, \method{} can be seamlessly integrated into generative pipelines as a replacement for traditional alignment procedures, producing morphings comparable to state-of-the-art generative approaches.

Despite these advantages, \method{} inherits certain limitations intrinsic to flow-based formulations.  
Since it enforces invertibility by construction, it cannot model transformations involving occlusions or topological changes, such as morphing between faces with open and closed mouths.  
In such cases, generative models can complement our approach during the blending stage to recover missing structures.  
Additionally, as in other warping-based techniques, \method{} depends on the quality and accuracy of landmark correspondences, whether extracted manually or through automated detectors.

As future work, we plan to extend \method{} to point cloud alignment and registration, and further improve our Gaussian morphing approach.

\section*{Acknowledgments}
Guilherme and Nuno would like to thank Fundação de Ciência e Tecnologia (FCT) projects UIDB/00048/2020\footnote{DOI: \url{https://doi.org/10.54499/UIDB/00048/2020}} and UIDP/00048/2020 for partially funding this work. Guilherme would also like to thank FCT project 2024.07681.IACDC\footnote{DOI: \url{https://doi.org/10.54499/2024.07681.IACDC}} for partially funding this work. João Paulo would like to thank Fundação Carlos Chagas Filho de Amparo à Pesquisa do Estado do Rio de Janeiro (FAPERJ) grant SEI-260003/012808/2024 for funding this work.
Vitor and Daniel gratefully acknowledge support from CAPES, grants 88887.842584/2023-00 and 88887.832821/2023-00, respectively for supporting this research.
We also thank Google for funding this research.

\bibliographystyle{plainnat}
\bibliography{references}

\begin{thebibliography}{60}
\providecommand{\natexlab}[1]{#1}
\providecommand{\url}[1]{\texttt{#1}}
\expandafter\ifx\csname urlstyle\endcsname\relax
  \providecommand{\doi}[1]{doi: #1}\else
  \providecommand{\doi}{doi: \begingroup \urlstyle{rm}\Url}\fi

\bibitem[Anokhin et~al.(2021)Anokhin, Demochkin, Khakhulin, Sterkin, Lempitsky, and Korzhenkov]{anokhin2021image}
Ivan Anokhin, Kirill Demochkin, Taras Khakhulin, Gleb Sterkin, Victor Lempitsky, and Denis Korzhenkov.
\newblock Image generators with conditionally-independent pixel synthesis.
\newblock In \emph{CVPR}, pages 14278--14287, 2021.

\bibitem[Balakrishnan et~al.(2019)Balakrishnan, Zhao, Sabuncu, Guttag, and Dalca]{balakrishnan2019voxelmorph}
Guha Balakrishnan, Amy Zhao, Mert~R Sabuncu, John Guttag, and Adrian~V Dalca.
\newblock Voxelmorph: a learning framework for deformable medical image registration.
\newblock \emph{IEEE transactions on medical imaging}, 2019.

\bibitem[Barron et~al.(2022)Barron, Mildenhall, Verbin, Srinivasan, and Hedman]{barron2022mipnerf}
Jonathan~T. Barron, Ben Mildenhall, Dor Verbin, Pratul~P. Srinivasan, and Peter Hedman.
\newblock Mip-nerf 360: Unbounded anti-aliased neural radiance fields.
\newblock \emph{CVPR}, 2022.

\bibitem[Beier and Neely(1992)]{beier1992feature}
Thaddeus Beier and Shawn Neely.
\newblock Feature-based image metamorphosis.
\newblock \emph{ACM SIGGRAPH computer graphics}, 26\penalty0 (2):\penalty0 35--42, 1992.

\bibitem[Bizzi et~al.(2025{\natexlab{a}})Bizzi, Moreira, Marques, Mendon{\c{c}}a, de~Oliveira, Balestro, Fernandez, Yukimura, Petrov, Pereira, Novello, and Lucas]{bizzi2025neuro}
Arthur Bizzi, Leonardo~M Moreira, M{\'a}rcio Marques, Leonardo Mendon{\c{c}}a, Christian~J{\'u}nior de~Oliveira, Vitor Balestro, Lucas dos~Santos Fernandez, Daniel Yukimura, Pavel Petrov, Jo{\~a}o Pereira, Tiago Novello, and Nissenbaum Lucas.
\newblock Neuro-spectral architectures for causal physics-informed networks.
\newblock \emph{arXiv preprint arXiv:2509.04966}, 2025{\natexlab{a}}.

\bibitem[Bizzi et~al.(2025{\natexlab{b}})Bizzi, Nissenbaum, and Pereira]{bizzi2024neural}
Arthur Bizzi, Lucas Nissenbaum, and João~M. Pereira.
\newblock Neural conjugate flows: A physics-informed architecture with flow structure.
\newblock \emph{Proceedings of the AAAI Conference on Artificial Intelligence}, 39\penalty0 (15):\penalty0 15576--15586, Apr. 2025{\natexlab{b}}.
\newblock \doi{10.1609/aaai.v39i15.33710}.
\newblock URL \url{https://ojs.aaai.org/index.php/AAAI/article/view/33710}.

\bibitem[Bookstein(1989)]{bookstein1989principal}
Fred~L. Bookstein.
\newblock Principal warps: Thin-plate splines and the decomposition of deformations.
\newblock \emph{IEEE Transactions on pattern analysis and machine intelligence}, 11\penalty0 (6):\penalty0 567--585, 1989.

\bibitem[Butler et~al.(2012)Butler, Wulff, Stanley, and Black]{butler2012naturalistic}
Daniel~J Butler, Jonas Wulff, Garrett~B Stanley, and Michael~J Black.
\newblock A naturalistic open source movie for optical flow evaluation.
\newblock In \emph{Computer Vision--ECCV 2012: 12th European Conference on Computer Vision, Florence, Italy, October 7-13, 2012, Proceedings, Part VI 12}, pages 611--625. Springer, 2012.

\bibitem[Chen et~al.(2018)Chen, Rubanova, Bettencourt, and Duvenaud]{chen2018neural}
Ricky~TQ Chen, Yulia Rubanova, Jesse Bettencourt, and David~K Duvenaud.
\newblock Neural ordinary differential equations.
\newblock \emph{Neurips}, 31, 2018.

\bibitem[DeBruine and Jones(2017)]{debruine2017frll}
Lisa DeBruine and Benedict Jones.
\newblock Face research lab london set, May 2017.

\bibitem[Dinh et~al.(2017)Dinh, Sohl-Dickstein, and Bengio]{dinh2017density}
Laurent Dinh, Jascha Sohl-Dickstein, and Samy Bengio.
\newblock Density estimation using real nvp.
\newblock In \emph{International Conference on Learning Representations}, 2017.

\bibitem[Gomes et~al.(1999)Gomes, Darsa, Costa, and Velho]{gomes1999warping}
Jonas Gomes, Lucia Darsa, Bruno Costa, and Luiz Velho.
\newblock \emph{Warping \& morphing of graphical objects}.
\newblock Morgan Kaufmann, 1999.

\bibitem[Grimmer and Busch(2024)]{grimmer2024ladimofacemorphgeneration}
Marcel Grimmer and Christoph Busch.
\newblock Ladimo: Face morph generation through biometric template inversion with latent diffusion, 2024.
\newblock URL \url{https://arxiv.org/abs/2410.07988}.

\bibitem[Gropp et~al.(2020)Gropp, Yariv, Haim, Atzmon, and Lipman]{gropp2020implicit}
Amos Gropp, Lior Yariv, Niv Haim, Matan Atzmon, and Yaron Lipman.
\newblock Implicit geometric regularization for learning shapes.
\newblock In \emph{37th International Conference on Machine Learning, ICML 2020}, pages 3747--3757, 2020.

\bibitem[He et~al.(2016)He, Zhang, Ren, and Sun]{he2016resnet}
Kaiming He, Xiangyu Zhang, Shaoqing Ren, and Jian Sun.
\newblock Deep residual learning for image recognition.
\newblock In \emph{2016 IEEE Conference on Computer Vision and Pattern Recognition (CVPR)}, pages 770--778, 2016.
\newblock \doi{10.1109/CVPR.2016.90}.

\bibitem[Heusel et~al.(2017)Heusel, Ramsauer, Unterthiner, Nessler, and Hochreiter]{heusel2017fid}
Martin Heusel, Hubert Ramsauer, Thomas Unterthiner, Bernhard Nessler, and Sepp Hochreiter.
\newblock {GANs} trained by a two time-scale update rule converge to a local nash equilibrium.
\newblock In I.~Guyon, U.~Von Luxburg, S.~Bengio, H.~Wallach, R.~Fergus, S.~Vishwanathan, and R.~Garnett, editors, \emph{Neurips}, volume~30. Curran Associates, Inc., 2017.
\newblock URL \url{https://proceedings.neurips.cc/paper_files/paper/2017/file/8a1d694707eb0fefe65871369074926d-Paper.pdf}.

\bibitem[Huang et~al.(2022)Huang, Zhang, Heng, Shi, and Zhou]{huang2022real}
Zhewei Huang, Tianyuan Zhang, Wen Heng, Boxin Shi, and Shuchang Zhou.
\newblock Real-time intermediate flow estimation for video frame interpolation.
\newblock In \emph{European Conference on Computer Vision}, pages 624--642. Springer, 2022.

\bibitem[Karras et~al.(2020)Karras, Laine, Aittala, Hellsten, Lehtinen, and Aila]{karras2020stylegan}
Tero Karras, Samuli Laine, Miika Aittala, Janne Hellsten, Jaakko Lehtinen, and Timo Aila.
\newblock Analyzing and improving the image quality of stylegan.
\newblock In \emph{2020 {IEEE/CVF} Conference on Computer Vision and Pattern Recognition}, pages 8107--8116. Computer Vision Foundation / {IEEE}, 2020.
\newblock \doi{10.1109/CVPR42600.2020.00813}.

\bibitem[Kazemi and Sullivan(2014)]{kazemi2014cvpr}
Vahid Kazemi and Josephine Sullivan.
\newblock One millisecond face alignment with an ensemble of regression trees.
\newblock In \emph{CVPR}, June 2014.

\bibitem[Kerbl et~al.(2023)Kerbl, Kopanas, Leimk{\"u}hler, and Drettakis]{kerbl3Dgaussians}
Bernhard Kerbl, Georgios Kopanas, Thomas Leimk{\"u}hler, and George Drettakis.
\newblock 3d gaussian splatting for real-time radiance field rendering.
\newblock \emph{ACM Transactions on Graphics}, 42\penalty0 (4), July 2023.
\newblock URL \url{https://repo-sam.inria.fr/fungraph/3d-gaussian-splatting/}.

\bibitem[Kingma and Ba(2017)]{kingma2014adam}
Diederik~P. Kingma and Jimmy Ba.
\newblock Adam: A method for stochastic optimization, 2017.
\newblock URL \url{https://arxiv.org/abs/1412.6980}.

\bibitem[Kirschstein et~al.(2023)Kirschstein, Qian, Giebenhain, Walter, and Nie{\ss}ner]{kirschstein2023nersemble}
Tobias Kirschstein, Shenhan Qian, Simon Giebenhain, Tim Walter, and Matthias Nie{\ss}ner.
\newblock Nersemble: Multi-view radiance field reconstruction of human heads.
\newblock \emph{ACM Transactions on Graphics (TOG)}, 42\penalty0 (4):\penalty0 1--14, 2023.

\bibitem[Kong et~al.(2022)Kong, Jiang, Luo, Chu, Huang, Tai, Wang, and Yang]{kong2022ifrnet}
Lingtong Kong, Boyuan Jiang, Donghao Luo, Wenqing Chu, Xiaoming Huang, Ying Tai, Chengjie Wang, and Jie Yang.
\newblock Ifrnet: Intermediate feature refine network for efficient frame interpolation.
\newblock In \emph{CVPR}, pages 1969--1978, 2022.

\bibitem[Li et~al.(2017)Li, Bolkart, Black, Li, and Romero]{li2017learning}
Tianye Li, Timo Bolkart, Michael~J Black, Hao Li, and Javier Romero.
\newblock Learning a model of facial shape and expression from 4d scans.
\newblock \emph{ACM Trans. Graph.}, 36\penalty0 (6):\penalty0 194--1, 2017.

\bibitem[Li and Snavely(2018)]{li2018megadepth}
Zhengqi Li and Noah Snavely.
\newblock Megadepth: Learning single-view depth prediction from internet photos.
\newblock In \emph{CVPR}, pages 2041--2050, 2018.

\bibitem[Liao et~al.(2014)Liao, Lima, Nehab, Hoppe, Sander, and Yu]{liao2014automating}
Jing Liao, Rodolfo~S Lima, Diego Nehab, Hugues Hoppe, Pedro~V Sander, and Jinhui Yu.
\newblock Automating image morphing using structural similarity on a halfway domain.
\newblock \emph{ACM Transactions on Graphics (TOG)}, 33\penalty0 (5):\penalty0 1--12, 2014.

\bibitem[Mehta et~al.(2022)Mehta, Chandraker, and Ramamoorthi]{ishit2022levelset}
Ishit Mehta, Manmohan Chandraker, and Ravi Ramamoorthi.
\newblock A level set theory for neural implicit evolution under explicit flows.
\newblock In \emph{Computer Vision -- ECCV 2022}, pages 711--729, Cham, 2022. Springer Nature Switzerland.
\newblock ISBN 978-3-031-20086-1.

\bibitem[Mildenhall et~al.(2020)Mildenhall, Srinivasan, Tancik, Barron, Ramamoorthi, and Ng]{mildenhall2020nerf}
Ben Mildenhall, Pratul~P Srinivasan, Matthew Tancik, Jonathan~T Barron, Ravi Ramamoorthi, and Ren Ng.
\newblock Nerf: Representing scenes as neural radiance fields for view synthesis.
\newblock In \emph{European Conference on Computer Vision}, pages 405--421. Springer, 2020.

\bibitem[Neidinger(2010)]{neidinger2010introduction}
Richard~D Neidinger.
\newblock Introduction to automatic differentiation and matlab object-oriented programming.
\newblock \emph{SIAM review}, 52\penalty0 (3):\penalty0 545--563, 2010.

\bibitem[Novello et~al.(2022)Novello, Schardong, Schirmer, Da~Silva, Lopes, and Velho]{novello2022exploring}
Tiago Novello, Guilherme Schardong, Luiz Schirmer, Vin{\'\i}cius Da~Silva, H{\'e}lio Lopes, and Luiz Velho.
\newblock Exploring differential geometry in neural implicits.
\newblock \emph{Computers \& Graphics}, 108:\penalty0 49--60, 2022.

\bibitem[Novello et~al.(2023)Novello, Da~Silva, Schardong, Schirmer, Lopes, and Velho]{novello2023neural}
Tiago Novello, Vinicius Da~Silva, Guilherme Schardong, Luiz Schirmer, Helio Lopes, and Luiz Velho.
\newblock Neural implicit surface evolution.
\newblock In \emph{Proceedings of the IEEE/CVF international conference on computer vision}, pages 14279--14289, 2023.

\bibitem[Novello et~al.(2025)Novello, Aldana, Araujo, and Velho]{novello2025tuningfrequenciesrobusttraining}
Tiago Novello, Diana Aldana, Andre Araujo, and Luiz Velho.
\newblock Tuning the frequencies: Robust training for sinusoidal neural networks.
\newblock In \emph{Proceedings of the Computer Vision and Pattern Recognition Conference}, pages 3071--3080, 2025.

\bibitem[Paszke et~al.(2019)Paszke, Gross, Massa, Lerer, Bradbury, Chanan, Killeen, Lin, Gimelshein, Antiga, Desmaison, Köpf, Yang, DeVito, Raison, Tejani, Chilamkurthy, Steiner, Fang, Bai, and Chintala]{paszke2019pytorch}
Adam Paszke, Sam Gross, Francisco Massa, Adam Lerer, James Bradbury, Gregory Chanan, Trevor Killeen, Zeming Lin, Natalia Gimelshein, Luca Antiga, Alban Desmaison, Andreas Köpf, Edward Yang, Zach DeVito, Martin Raison, Alykhan Tejani, Sasank Chilamkurthy, Benoit Steiner, Lu~Fang, Junjie Bai, and Soumith Chintala.
\newblock Pytorch: An imperative style, high-performance deep learning library, 2019.
\newblock URL \url{https://arxiv.org/abs/1912.01703}.

\bibitem[Paz et~al.(2023)Paz, Perazzo, Novello, Schardong, Schirmer, da~Silva, Yukimura, Chagas, Lopes, and Velho]{paz2023mr}
Hallison Paz, Daniel Perazzo, Tiago Novello, Guilherme Schardong, Luiz Schirmer, Vinicius da~Silva, Daniel Yukimura, Fabio Chagas, Helio Lopes, and Luiz Velho.
\newblock Mr-net: Multiresolution sinusoidal neural networks.
\newblock \emph{Computers \& Graphics}, 2023.

\bibitem[Paz et~al.(2024)Paz, Novello, and Velho]{paz2024spectral}
Hallison Paz, Tiago Novello, and Luiz Velho.
\newblock Spectral periodic networks for neural rendering.
\newblock In \emph{ACM SIGGRAPH 2024 Posters}, pages 1--2. 2024.

\bibitem[Pereira~Matias et~al.(2025)Pereira~Matias, Perazzo, Silva, Raposo, Velho, Paiva, and Novello]{matias2025gaussian}
Vitor Pereira~Matias, Daniel Perazzo, Vinicius Silva, Alberto Raposo, Luiz Velho, Afonso Paiva, and Tiago Novello.
\newblock From volume rendering to 3d gaussian splatting: Theory and applications.
\newblock In \emph{SIBGRAPI Conference on Graphics, Patterns and Images}, 2025.

\bibitem[P{\'e}rez et~al.(2023)P{\'e}rez, Gangnet, and Blake]{perez2023poisson}
Patrick P{\'e}rez, Michel Gangnet, and Andrew Blake.
\newblock Poisson image editing.
\newblock In \emph{Seminal Graphics Papers: Pushing the Boundaries, Volume 2}, pages 577--582. 2023.

\bibitem[Potje et~al.(2024)Potje, Cadar, Araujo, Martins, and Nascimento]{potje2024xfeat}
Guilherme Potje, Felipe Cadar, Andr{\'e} Araujo, Renato Martins, and Erickson~R Nascimento.
\newblock Xfeat: Accelerated features for lightweight image matching.
\newblock In \emph{CVPR}, pages 2682--2691, 2024.

\bibitem[Qian et~al.(2024)Qian, Kirschstein, Schoneveld, Davoli, Giebenhain, and Nießner]{qian2024gaussianavatarsphotorealisticheadavatars}
Shenhan Qian, Tobias Kirschstein, Liam Schoneveld, Davide Davoli, Simon Giebenhain, and Matthias Nießner.
\newblock {GaussianAvatars}: Photorealistic head avatars with rigged 3d gaussians, 2024.
\newblock URL \url{https://arxiv.org/abs/2312.02069}.

\bibitem[Sagonas et~al.(2016)Sagonas, Antonakos, Tzimiropoulos, Zafeiriou, and Pantic]{sagonas2016}
Christos Sagonas, Epameinondas Antonakos, Georgios Tzimiropoulos, Stefanos Zafeiriou, and Maja Pantic.
\newblock 300 faces in-the-wild challenge: database and results.
\newblock \emph{Image and Vision Computing}, 47:\penalty0 3--18, 2016.
\newblock ISSN 0262-8856.
\newblock \doi{https://doi.org/10.1016/j.imavis.2016.01.002}.
\newblock URL \url{https://www.sciencedirect.com/science/article/pii/S0262885616000147}.
\newblock 300-W, the First Automatic Facial Landmark Detection in-the-Wild Challenge.

\bibitem[Sang et~al.(2025{\natexlab{a}})Sang, Canfes, Cao, Bernard, and Cremers]{sang2025implicit}
Lu~Sang, Zehranaz Canfes, Dongliang Cao, Florian Bernard, and Daniel Cremers.
\newblock Implicit neural surface deformation with explicit velocity fields, 2025{\natexlab{a}}.
\newblock URL \url{https://arxiv.org/abs/2501.14038}.

\bibitem[Sang et~al.(2025{\natexlab{b}})Sang, Canfes, Cao, Marin, Bernard, and Cremers]{sang20254deform}
Lu~Sang, Zehranaz Canfes, Dongliang Cao, Riccardo Marin, Florian Bernard, and Daniel Cremers.
\newblock 4deform: Neural surface deformation for robust shape interpolation.
\newblock In \emph{Proceedings of the Computer Vision and Pattern Recognition Conference}, 2025{\natexlab{b}}.

\bibitem[Sarkar et~al.(2020)Sarkar, Korshunov, Colbois, and Marcel]{sarkar2020landmarks}
Eklavya Sarkar, Pavel Korshunov, Laurent Colbois, and S\'{e}bastien Marcel.
\newblock Vulnerability analysis of face morphing attacks from landmarks and generative adversarial networks, October 2020.
\newblock URL \url{https://arxiv.org/abs/2012.05344}.

\bibitem[Schardong et~al.(2024)Schardong, Novello, Paz, Medvedev, Da~Silva, Velho, and Gon{\c{c}}alves]{schardong2024neural}
Guilherme Schardong, Tiago Novello, Hallison Paz, Iurii Medvedev, Vin{\'\i}cius Da~Silva, Luiz Velho, and Nuno Gon{\c{c}}alves.
\newblock Neural implicit morphing of face images.
\newblock In \emph{Proceedings of the IEEE/CVF Conference on Computer Vision and Pattern Recognition}, pages 7321--7330, 2024.

\bibitem[Schirmer et~al.(2024)Schirmer, Novello, da~Silva, Schardong, Perazzo, Lopes, Gon{\c{c}}alves, and Velho]{schirmer2024geometric}
Luiz Schirmer, Tiago Novello, Vin{\'\i}cius da~Silva, Guilherme Schardong, Daniel Perazzo, H{\'e}lio Lopes, Nuno Gon{\c{c}}alves, and Luiz Velho.
\newblock Geometric implicit neural representations for signed distance functions.
\newblock \emph{Computers \& Graphics}, 125:\penalty0 104085, 2024.

\bibitem[Singh and Ramachandra(2024)]{singh2024facemorphing}
Jag~Mohan Singh and Raghavendra Ramachandra.
\newblock 3-d face morphing attacks: Generation, vulnerability and detection.
\newblock \emph{IEEE Transactions on Biometrics, Behavior, and Identity Science}, 6\penalty0 (1):\penalty0 103--117, 2024.
\newblock \doi{10.1109/TBIOM.2023.3324684}.

\bibitem[Sitzmann et~al.(2020)Sitzmann, Martel, Bergman, Lindell, and Wetzstein]{sitzmann2020implicit}
Vincent Sitzmann, Julien Martel, Alexander Bergman, David Lindell, and Gordon Wetzstein.
\newblock Implicit neural representations with periodic activation functions.
\newblock \emph{Neurips}, 33, 2020.

\bibitem[Smythe(1990)]{smythe1990meshwarp}
Douglas Smythe.
\newblock A two-pass mesh warping algorithm for object transformation and image interpolation.
\newblock Technical Report 1030, ILM Technical Memo, Computer Graphics Department, Lucasfilm Ltd., 1990.

\bibitem[Sorkine and Alexa(2007)]{olga2007arap}
Olga Sorkine and Marc Alexa.
\newblock As-rigid-as-possible surface modeling.
\newblock In \emph{Proceedings of the Fifth Eurographics Symposium on Geometry Processing}, SGP '07, page 109–116, Goslar, DEU, 2007. Eurographics Association.
\newblock ISBN 9783905673463.

\bibitem[Sun et~al.(2022)Sun, Han, Kong, Tang, Yan, and Xie]{sun2022topology}
Shanlin Sun, Kun Han, Deying Kong, Hao Tang, Xiangyi Yan, and Xiaohui Xie.
\newblock Topology-preserving shape reconstruction and registration via neural diffeomorphic flow.
\newblock In \emph{CVPR}, 2022.

\bibitem[Sun et~al.(2024)Sun, Han, You, Tang, Kong, Naushad, Yan, Ma, Khosravi, Duncan, et~al.]{sun2024medical}
Shanlin Sun, Kun Han, Chenyu You, Hao Tang, Deying Kong, Junayed Naushad, Xiangyi Yan, Haoyu Ma, Pooya Khosravi, James~S Duncan, et~al.
\newblock Medical image registration via neural fields.
\newblock \emph{Medical Image Analysis}, 97:\penalty0 103249, 2024.

\bibitem[Tancik et~al.(2020)Tancik, Srinivasan, Mildenhall, Fridovich-Keil, Raghavan, Singhal, Ramamoorthi, Barron, and Ng]{tancik2020fourier}
Matthew Tancik, Pratul Srinivasan, Ben Mildenhall, Sara Fridovich-Keil, Nithin Raghavan, Utkarsh Singhal, Ravi Ramamoorthi, Jonathan Barron, and Ren Ng.
\newblock Fourier features let networks learn high frequency functions in low dimensional domains.
\newblock \emph{Neurips}, 33:\penalty0 7537--7547, 2020.

\bibitem[Teed and Deng(2020)]{teed2020raft}
Zachary Teed and Jia Deng.
\newblock Raft: Recurrent all-pairs field transforms for optical flow.
\newblock In \emph{Computer Vision--ECCV 2020: 16th European Conference, Glasgow, UK, August 23--28, 2020, Proceedings, Part II 16}, pages 402--419. Springer, 2020.

\bibitem[Viana and Espinar(2021)]{viana2021differential}
Marcelo Viana and Jos{\'e}~M Espinar.
\newblock \emph{Differential equations: a dynamical systems approach to theory and practice}, volume 212.
\newblock American Mathematical Society, 2021.

\bibitem[Wolberg(1990)]{wolberg1990digital}
George Wolberg.
\newblock \emph{Digital image warping}, volume 10662.
\newblock IEEE computer society press Los Alamitos, CA, 1990.

\bibitem[Wolberg(1998)]{wolberg1998image}
George Wolberg.
\newblock Image morphing: a survey.
\newblock \emph{The visual computer}, 14\penalty0 (8-9):\penalty0 360--372, 1998.

\bibitem[Wu et~al.(2022)Wu, Jiahao, Wang, Yushkevich, Hsieh, and Gee]{wu2022nodeo}
Yifan Wu, Tom~Z Jiahao, Jiancong Wang, Paul~A Yushkevich, M~Ani Hsieh, and James~C Gee.
\newblock Nodeo: A neural ordinary differential equation based optimization framework for deformable image registration.
\newblock In \emph{CVPR}, 2022.

\bibitem[Yang et~al.(2021)Yang, Belongie, Hariharan, and Koltun]{yang2021geometry}
Guandao Yang, Serge Belongie, Bharath Hariharan, and Vladlen Koltun.
\newblock Geometry processing with neural fields.
\newblock \emph{Neurips}, 34:\penalty0 22483--22497, 2021.

\bibitem[Zhang et~al.(2024)Zhang, Zhou, Xu, Dai, and Pan]{zhang2024diffmorpher}
Kaiwen Zhang, Yifan Zhou, Xudong Xu, Bo~Dai, and Xingang Pan.
\newblock Diffmorpher: Unleashing the capability of diffusion models for image morphing.
\newblock In \emph{CVPR}, pages 7912--7921, 2024.

\bibitem[Zhang et~al.(2018)Zhang, Isola, Efros, Shechtman, and Wang]{zhang2018perceptual}
Richard Zhang, Phillip Isola, Alexei~A Efros, Eli Shechtman, and Oliver Wang.
\newblock The unreasonable effectiveness of deep features as a perceptual metric.
\newblock In \emph{CVPR}, 2018.

\end{thebibliography}

\appendix

\section{Forward differentiation}
\label{a-FD}

We implement forward differentiation (FD) to speed up the calculation of Jacobian or Hessian terms, for thin-plate regularization.
Forward mode differentiation is an alternative way to implement automatic differentiation, opposite to backward (or reverse mode) differentiation.
FD is typically implemented using tangent (or dual) numbers \cite{neidinger2010introduction}. Let us illustrate FD with an example. Suppose we want to evaluate $\frac{\partial x_n}{\partial x_0}$, where
\begin{equation}\label{eq:recursion}
x_i = f_i(x_{i-1}),\quad i=1,\dots,n.
\end{equation}
Using the chain rule, we obtain that 
\begin{equation*}
\frac{\partial x_n}{\partial x_0} = \prod_{i=1}^{n} \frac{\partial x_i}{\partial x_{i-1}} = \prod_{i=1}^{n}  f_i'(x_{i-1}).
\end{equation*}
To implement FD, we introduce the tangent (or dual) variables $\dot x_k$, $k=0,\dots,n$, defined by the recursion $\dot x_0 = \frac{\partial x_0}{\partial x_0}=1$, and
\begin{equation}\label{eq:fd_recursion}
\dot x_i = f'_i(x_{i-1}) \dot x_{i-1},\quad i=1,\dots,n
\end{equation}
Using induction, it follows that
\begin{equation*}
\dot x_k = f_k'(x_{k-1}) \dot x_{k-1} = \prod_{i=1}^{k} f_i'(x_{i-1}) = \frac{\partial x_k}{\partial x_0},\quad k=1,\dots,n.
\end{equation*}
Finally, the last element yields the desired derivative $\dot x_n=\frac{\partial x_n}{\partial x_0}$. To implement this approach, we can replace every function $f_i$ by an FD version, that calculates both $x_{i}$ and $\dot x_{i}$ at the same time. Effectively, the FD version of $f_i$ maps ordered pairs $\fdord{x_{i-1}, \dot x_{i-1}}$ to ordered pairs $\fdord{x_{i}, \dot x_{i}}$, so that both \eqref{eq:recursion} and \eqref{eq:fd_recursion} are satisfied:
\begin{equation*}
\fdmap{x_{i}, \dot x_{i}} = f_i^{\text{FD}}\fdmap{x_{i-1}, \dot x_{i-1}} := \fdord{f_i(x_{i-1}), f'_i(x_{i-1}) \dot x_{i-1}},\quad i=1,\dots,n
\end{equation*}%
%, implementation this approach, the function arguments are replaced by ordered pairs, $\left\langle y, \dot{y} \right\rangle$, where $\dot{y}=\frac{\partial y}{\partial x}$. Correspondingly, functions map ordered pairs to ordered pairs as to preserve the chain rule. For instance, if $g:\R\to \R$, we have
%\begin{equation}
%g_{\text{FD}}(\left\langle y, \dot{y} \right\rangle) = \left\langle g(y), \frac{\partial }{\partial x}\left(g(y)\right)\right\rangle = \left\langle g(y), g'(y) \frac{\partial y}{\partial x} \right\rangle = \left\langle g(y), g'(y)\dot{y} \right\rangle
%\end{equation}

Since we need to calculate Hessians, we develop a multivariate FD implementation involving ordered triplets $\fdord{z, \dot{z}, \ddot{z}}$. Here, $\dot{z}$ contains intermediary gradients, and $\ddot{z}$ contains intermediary Hessians, with respect to a variable $x$. More specifically, if $z\in \R^m$ and $x\in \R^n$, our implementation ensures that $\dot{z}=\grad[x]z \in \R^{m\times n}$ and $\ddot{z}=\hess[x]{z}\in\R^{m\times n \times n}$.

To calculate the Hessian of the model $f_\theta(x)$ with respect to $x$, we first write it as a composition of simpler functions (linear operators, non-linear activations, flow operator of affine flows), for which we have implemented forward differentiation: $f_\theta = f_{1}\circ f_{2}\circ \cdots \circ f_{m}$. Then, noting that $\grad[x]x = \mathbf{I}_n$ (the $n\times n$ identity matrix) and $\hess[x]{x}= \mathbf{0}_{n\times n\times n}$, we calculate 
\begin{equation*}
f^{\text{FD}}_\theta\fdmap{x, \mathbf{I}_n, \mathbf{0}_{n\times n\times n}} = f_{1}^{\text{FD}}\circ \cdots \circ f_{m}^{\text{FD}}\fdmap{x, \mathbf{I}_n, \mathbf{0}_{n\times n\times n}} = \fdord{u, \dot u , \ddot u},
\end{equation*}
and set $\hess[x]{f_\theta(x)}=\ddot u$.

Followingly, we provide the details for the FD implementation of linear transformations, non-linear activations, and the flow operator of an affine flow. By composing these, we can construct all architectures we consider in the paper.
\paragraph{Linear operators:} Consider the linear operator $\mathcal{A}\mathbf{z}\mapsto \mathbf{A}\mathbf{z} + \mathbf{b}$. Since derivatives are also linear operators, the FD implementation is straightforward.
\begin{equation*}
\mathcal{A}^{\text{FD}}\fdord{\mathbf{z},\mathbf{\dot z}, \mathbf{\ddot z}} = \fdord{\mathbf{A}\mathbf{z} + \mathbf{b},\, \mathbf{A}\mathbf{\dot z},\, \mathbf{A}\times_1 \mathbf{\ddot z}},
\end{equation*}
where $\mathbf{A}\times_1$ denotes multiplication by $\mathbf{A}$ in the first dimension.
\paragraph{Non-linear activations:} Suppose that $z\in \R$, $g:\R\to \R$, $\dot z \in \R^n$ and $\ddot z\in \R^{n\times n}$. Then using the chain rule, one obtains:
\begin{align*}
g^{\text{FD}}\fdmap{z, \dot z, \ddot z} &= \fdord{g(z),\, \grad{g(z)},\, \hess{g(z)}},\\
&= \fdord{g(z),\, g'(z)\dot z,\, \grad{(g'(z)\dot z)}},\\
&= \fdord{g(z),\, g'(z)\dot z,\, g'(z)\ddot z + g''(z)\dot z \dot{z}^T}.
\end{align*}
Here, the terms $g(z), g'(z)$ and $g''(z)$ are the derivatives of the activation function. For instance, for SIRENs, we have $g(z)=\frac{1}{w_0} \sin(w_0 z)$, $g'(z)= \cos(w_0 z)$ and $g''(z)=-w_0 \sin(w_0 z)=-w_0^2 g(z)$.

\paragraph{Flow operator of an Affine Flow:}
The formula for an affine flow is provided in \cite{bizzi2024neural}.
\begin{equation*}
    \Phi(\x, t) = e^{\mathbf{A} t} \x + \int_{0}^t e^{\mathbf{A} \tau} \mathbf{b}\,d\tau,
\end{equation*}
where the affine integral term is calculated exactly using an augmentation trick \cite{bizzi2024neural}. This flow has some nice properties: the map is linear on $\x$ and the derivatives of the exponential in terms of $t$ are easier to calculate. Suppose we want to evaluate $\Phi_{\text{FD}}(\left\langle \x, \dot{\x}, \ddot{\x}\right\rangle, \left\langle t, \dot t, \ddot t\right\rangle)$, where $\x\in \R^d$, $\dot{\x} \in \R^{d\times n}$, $\ddot{\x}\in \R^{d\times n\times n}$, $t\in \R$, $\dot t \in \R^{n}$ and $\ddot t\in \R^{n\times n}$. Letting $\mathbf{f} = \Phi(\x, t)$, $\dot{\mathbf{f}} = \mathbf{A}\mathbf{f}+\mathbf{b}$ and $\mathbf{E} =  e^{\mathbf{A} t}$, we obtain, using the chain rule:
\begin{align*}
\Phi^{\text{FD}}\fdmap{\fdord{\x, \dot{\x}, \ddot{\x}},\fdord{t, \dot t, \ddot t}} &= \fdord{\Phi(\x, t), \,\nabla_x\Phi \,\dot \x + \Phi_t \dot t^T, \,\nabla^2_x \Phi\,\ddot \x + \nabla_x\Phi_t \otimes \dot t + \Phi_{tt} \otimes \ddot{t} },\\
&= \fdord{\mathbf{f}, \, \mathbf{E}\dot{\x} + \dot{\mathbf{f}} \dot t^T,  \mathbf{E}\times_1\ddot{\x}  + (\mathbf{A}\mathbf{E}\dot{\x})\otimes \dot{t} + (\mathbf{A} \dot{\mathbf{f}}) \otimes \ddot{t}}.
\end{align*}

\section{Ablation studies}
\label{a-ablation}
\paragraph{NODE integration steps.}
Neural ODEs require solving differential equations through numerical integration, both during training and inference, to associate each point with its corresponding source and target coordinates by integrating forward and backward in time.
Thus, it might appear that the trajectories need to be approximated using a large number of steps, making the process computationally expensive. However, we find that this is not the case: high-quality solutions can be obtained with only a few steps.
We sampled several initial conditions from the $[-1,1]^2$ grid and displaced the points using the trained NODE model with the same number of integration steps used during training. We then approximated the reference (ground-truth) trajectories by computing a high-resolution baseline with 1000 integration steps. As a quality metric, we measured the maximum squared error between the predicted trajectory points and the linearly interpolated baseline points; a lower error indicates that the numerical integration used during training accurately captures the underlying dynamics.
Table~\ref{tab:max_squared_error_step_ablation} shows the results of this experiment. We observe that only a small number of integration steps is sufficient to approximate the flow with high fidelity. Moreover, applying Jacobian regularization during training further stabilizes the dynamics, enabling reliable results even with very few integration steps.
\begin{table}[h] \centering \caption{Comparison of trajectory quality (maximum squared error) of NODEs with different integration steps against a baseline of 1000 steps. NODE-based morphing needs very few steps in order to obtain high quality approximations of the associated flow.} \label{tab:max_squared_error_step_ablation}  \begin{tabular}{ccccc} \toprule Int. steps & \multicolumn{2}{c}{MegaDepth Sample} & \multicolumn{2}{c}{FRLL Sample} \\ & No regularization & With regularization & No regularization & With regularization \\ \midrule 
3               &          5.44E-05 &            6.43E-07 &          1.36E-07 &            7.16E-10 \\
5               &          3.39E-08 &            3.27E-09 &          2.08E-10 &            6.19E-11 \\
7               &          1.88E-09 &            5.24E-10 &          5.46E-11 &            6.00E-11 \\
15              &          8.88E-12 &            5.18E-10 &          5.46E-11 &            6.10E-11 \\
\bottomrule \end{tabular} \end{table}

\begin{wrapfigure}[22]{R}{0.45\textwidth}
    \centering
    \vspace{-0.2cm}
    \includegraphics[width=\linewidth]{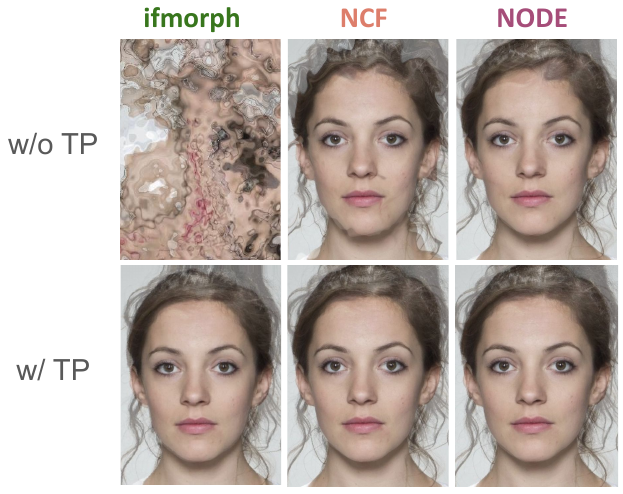}
    \vspace{-0.4cm}
\caption{Effect of TP regularization on {ifmorph} (left) and our methods, NCF (center) and NODE (right). The top row shows results without TP regularization, while the bottom row shows the same models with TP applied. TP regularization promotes smoother deformations and enhances structural coherence across the morphing process.}
    \label{fig:tps-ablation}
\end{wrapfigure}
\paragraph{Forward differentiation (FD).}
We conducted an experiment to evaluate the computational speed-up achieved by our FD scheme for Hessian computation. An NCF model was initialized using the same configuration as in our main experiments. For this model, we computed the Hessian at 1,000 domain points using both PyTorch’s \texttt{autograd} and our FD implementation, then evaluated the thin-plate loss and performed backpropagation.
Each method was run 100 times, and we report the average computation time per iteration. On an NVIDIA GeForce RTX~4090 GPU, our FD implementation achieved an average iteration time of $6.93 \times 10^{-3}$ seconds, compared to $1.62 \times 10^{-1}$ seconds using \texttt{autograd}, corresponding to a $23.4\times$ speed-up. This demonstrates the substantial efficiency gains of our FD approach over standard automatic differentiation.

\paragraph{Thin-plate regularization.}
We perform an ablation study to evaluate the effect of thin-plate (TP) regularization on deformation smoothness and structural consistency. As illustrated in \autoref{fig:tps-ablation}, TP regularization significantly improves the coherence of the deformation field, reducing distortions and enforcing smoother transitions between source and target landmarks.
The effect is most pronounced for the baseline {ifmorph}, which exhibits severe artifacts in the absence of regularization. In contrast, our models (NCF and NODE) already demonstrate stable deformation behavior due to their flow-based formulation, yet still benefit from TP regularization, yielding even more consistent and visually coherent morphings.

\section{Additional 3DGS morphing experiments}
\label{a-3dgs}

\paragraph{Flow evaluation}
Figure~\ref{fig:3d_flow_comparison} illustrates the 3D flow streamlines corresponding to landmark trajectories from the source image for three methods used in 3DGS morphing: {ifmorph}, and \method{} with NCF and NODE backbones.  
As shown, {ifmorph} produces flow fields that are less stable and exhibit pronounced curvature, while NCF and NODE yield smoother, more coherent, and regularized flows.  
This difference is particularly evident in regions not directly constrained by landmarks—such as the hair—where {ifmorph} introduces noticeable distortions.  
In contrast, the FLOWING variants preserve both structural continuity and spatial consistency across the field.  

\begin{figure}[ht!]
    \centering
    \includegraphics[width=\linewidth]{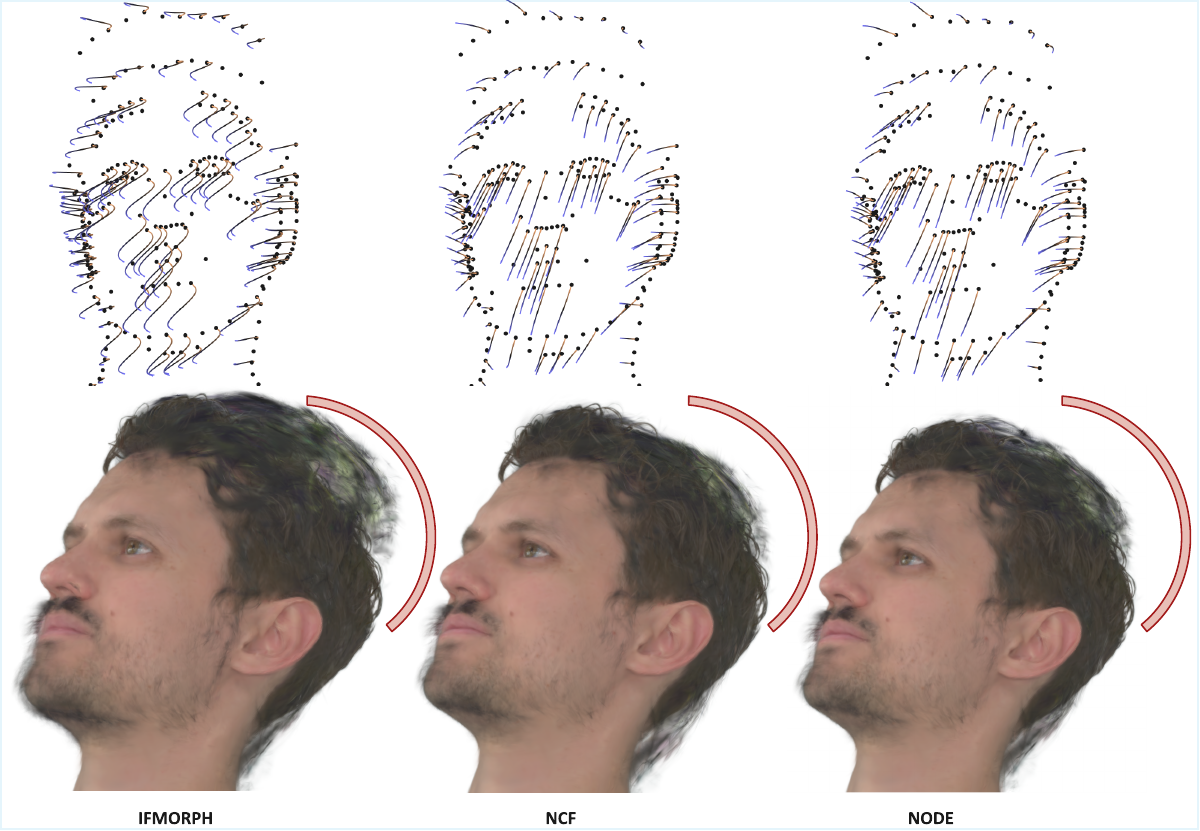}
    \caption{Visualization of 3D flow fields generated by {ifmorph}, NCF, and NODE in the 3DGS morphing task. The red arch highlights instability in the hair region for {ifmorph}, whereas NCF and NODE preserve structure more reliably.}
    \label{fig:3d_flow_comparison}
\end{figure}

\paragraph{Training details}
Our 3DGS morphing configuration extends the 2D morphing pipeline while maintaining the same loss weights.  
To accommodate the higher dimensionality, we use slightly larger learning rates (LRs) and an LR scheduler for both NODE and NCF to ensure convergence within 20{,}000 steps.  
The initial LRs are set to 0.001 for NODE, 0.002 for NCF, and 0.0001 for {ifmorph}.  
Early stopping is employed with a patience of 500 epochs for NCF and NODE, and 1{,}000 for {ifmorph}.  
If the loss plateaus for 100 epochs, the LR is reduced, and training terminates once the patience threshold is reached.  
The best-performing model is selected based on the lowest loss.

\paragraph{Additional results}
Figure~\ref{fig:3d_more_timesteps_comparison} presents additional qualitative results for our Gaussian morphing framework.  
\method{} consistently achieves smooth and coherent transitions across diverse subjects, effectively handling challenging regions such as facial and head hair (third row).  
Furthermore, it demonstrates strong generalization when morphing between subjects of different genders, maintaining both geometric structure and photometric consistency.  

\begin{figure}[ht!]
    \centering
    \includegraphics[width=\linewidth]{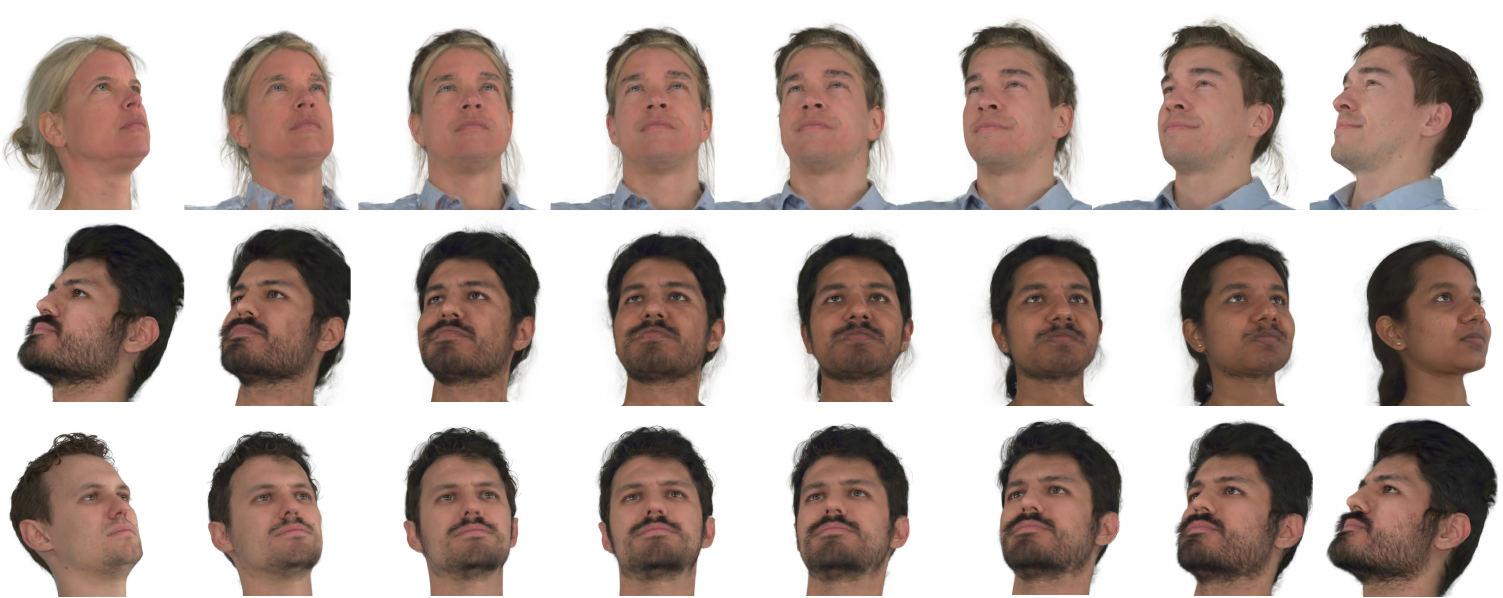}
    \caption{Additional results of Gaussian morphing across multiple timesteps (first and last columns show the targets). \method{} achieves smooth and coherent transitions across diverse subjects while preserving structure and appearance.}
    \label{fig:3d_more_timesteps_comparison}
\end{figure}

\section{Additional comparisons using generative blending} 
\label{a-generative}

Figure~\ref{fig:more-diffmorpher-examples} shows additional qualitative results combining \method{} with DiffMorpher as a generative blending strategy.  
These examples further confirm that introducing a warping stage prior to generative blending significantly improves the quality of the final morphs.  
In particular, \method{} enhances spatial alignment and structural coherence, producing smoother intermediate representations, even in simple scenarios such as morphing between spherical objects.
\begin{figure}[ht!]
    \centering
    \includegraphics[width=\linewidth]{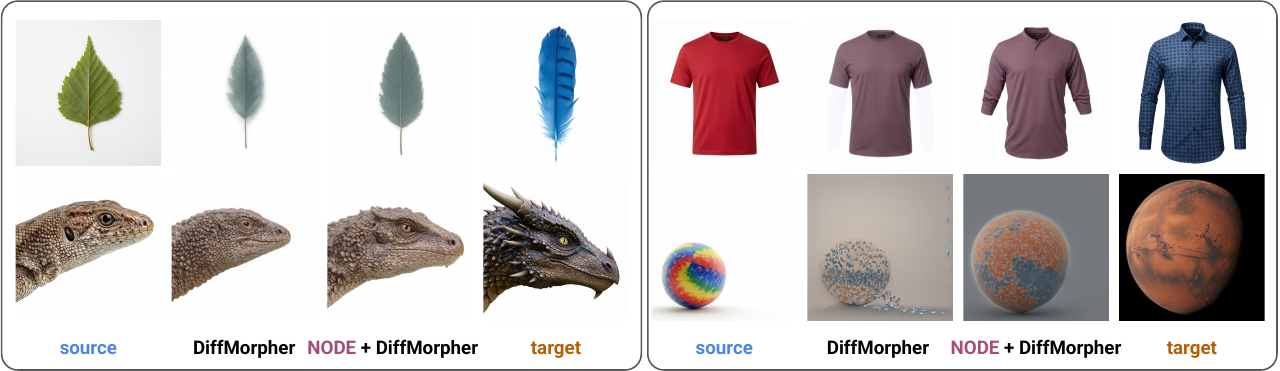}
    \caption{Additional examples comparing generative blending with and without \method{} as a warping prior. Incorporating \method{} before applying DiffMorpher (NODE+DiffMorpher, ours) yields superior structural preservation and smoother transitions.}
    \label{fig:more-diffmorpher-examples}
\end{figure}

\section{Applications in video interpolation} 
% \joaopaulo{we are not citing this appendix in the main text, should we do that?}
\label{a-video}

Video interpolation aims to synthesize intermediate frames between two given frames, a key component in applications such as video compression and frame rate upscaling.  
Most deep learning-based approaches rely on estimating optical flow between consecutive frames, which describes pixel motion as a vector field~\cite{huang2022real, kong2022ifrnet}.  
Assuming temporal proximity between frames, optical flow can be leveraged to approximate intermediate frames by integrating a time-independent ODE defined by the flow field.  

Inspired by this principle, we explore the use of \method{} to learn a compact representation of optical flow from sparse samples and to reconstruct intermediate frames.  
For this experiment, we employ a cropped sequence from the Sintel dataset~\cite{butler2012naturalistic} and use RAFT~\cite{teed2020raft} to generate dense optical flow between frames.  
We then sample a sparse set of image points and construct landmark pairs $(p^0, p^1)$ by displacing them according to the optical flow.  
\method{} is trained to model a continuous flow field aligning these points over time.
To evaluate interpolation quality, we warp and linearly blend the input frames at $t=0.5$ using the learned flow and compare the resulting image against the ground-truth middle frame.  
We further compare against three baselines: (1) linear frame blending, (2) dense optical-flow warping followed by linear blending, and (3) ifmorph~\cite{schardong2024neural}.  
As shown in Table~\ref{tab:optical_flow}, \method{} achieves lower MSE than all baseline interpolation methods, indicating improved accuracy in predicting intermediate frames.  
These preliminary results suggest that learning a continuous optical flow representation from sparse correspondences is a promising direction for video interpolation.  
% Future work includes extending this approach to larger and more diverse video datasets to assess generalization and robustness.
\begin{table}[h]
    \centering
\caption{Comparison of interpolation accuracy against the ground-truth intermediate frame on Sintel~\cite{butler2012naturalistic}. Lower MSE indicates better reconstruction.}
\label{tab:optical_flow}
\begin{tabular}{lc}
\toprule
{Method} & {MSE}~($\downarrow$) \\
\midrule
Linear Blend      &  2.6E\text{-}4 \\
Optical Flow Warp &  4.9E\text{-}5 \\ 
ifmorph           &  5.8E\text{-}5 \\
NCF (Ours)        &  \textbf{4.0E\text{-}5} \\
NODE (Ours)       &  \textbf{3.9E\text{-}5} \\
\bottomrule
\end{tabular} 
\end{table}

% \matias{Should this be here? Dunno where to put it, asked for the rebuttal}

\section{Comparison with TPS warping} 
\label{a-tps}
Another standard morphing algorithm is thin-plate-spline warping. While TPS warping does not involve an iterative process, the method does require solving a linear system of equations in order to find the appropriate linear combination of the basis functions. The matrix of such a system can, in some cases, be badly conditioned. When using the scikit-image implementation of the algorithm, we find that this problem leads to crashes or low-quality warpings in approximately 10\% of FRLL samples. This can negatively impact metrics such as the LPIPS. \method{}, on the other hand, does not present these problems as it does not rely on solving a similar badly-conditioned system. For comparison, using TPS warping combined with linear blending for the FRLL dataset, we obtain LPIPS$(I^0,I) = 0.371$ and LPIPS$(I^1,I)=0.374$.
\end{document}